\documentclass[sigconf]{acmart}

\usepackage{pifont}
\usepackage{enumitem}
\usepackage{algorithm}
\usepackage{algorithmic}
\usepackage{multirow}
\usepackage{booktabs}      
\usepackage{threeparttable}

\usepackage{tikz}
\usetikzlibrary{
    arrows.meta,
    positioning,
    calc,
    fit,
    shapes,
    shapes.geometric
}

\newcommand{\INPUT}{\item[\textbf{Input:}]}
\newcommand{\OUTPUT}{\item[\textbf{Output:}]}

\AtBeginDocument{%
  }

\acmISBN{978-1-4503-XXXX-X/2018/06}

\begin{document}

\settopmatter{printacmref=false}
\setcopyright{none}
\renewcommand\footnotetextcopyrightpermission[1]{}
\pagestyle{plain}

\title[FedEHR-Gen]{FedEHR-Gen: Federated Synthetic Time-Series EHR Generation via Latent Space Alignment and Distribution-Aware Aggregation}


\author{Jun Bai}
\affiliation{%
  \institution{McGill University}
  \institution{Mila - Quebec AI Institute}
  \city{Montreal}
  \country{Canada}
}

\author{Ziyang Song}
\affiliation{%
  \institution{Ohio University}
  \city{Athens}
  \country{United States}}

\author{Yue Li}
\authornote{Corresponding author:  Yue Li (yueli@cs.mcgill.ca)
}
\affiliation{%
  \institution{McGill University}
  \institution{Mila - Quebec AI Institute}
  \city{Montreal}
  \country{Canada}}









\begin{abstract}
Synthetic Electronic Health Record (EHR) generation provides a promising avenue for data augmentation and cross-hospital modeling in privacy-constrained healthcare settings.  However, most existing EHR generative models are centralized and require pooling data  across hospitals, which is often infeasible when real-world data sharing is restricted. While federated EHR generation offers a natural solution, direct federated modeling often collapses or diverges due to the high dimensionality, sparsity, and cross-hospital heterogeneity of EHR data. In this work, we propose FedEHR-Gen, the first federated framework for synthetic time-series EHR generation across distributed hospitals.
FedEHR-Gen uses a two-stage learning paradigm. First, we introduce a federated autoencoder that projects high-dimensional and sparse EHR features onto a compact latent space. To ensure semantic consistency across hospitals, we develop a layer-wise matching aggregation mechanism that aligns local encoders into a unified global latent space. 
Second, operating on this aligned latent space, we train a federated temporal conditional variational autoencoder (TCVAE) with distribution-aware aggregation, enabling stable temporal generative modeling under severe cross-hospital heterogeneity.
Extensive experiments on the eICU and MIMIC-III datasets demonstrate that FedEHR-Gen achieves generation fidelity, downstream utility, and privacy risk comparable to centralized training, while consistently outperforming the standard federated baseline.
\end{abstract}





\maketitle

\section{Introduction} \label{sec:introduction}

Time-series Electronic Health Records (TS-EHRs) capture temporal evolution of patient states and enable data-driven healthcare, including clinical risk analysis and  health trajectory modeling~\cite{park2025enhancing, wu2025harnessing}. EHRs contain protected health information and are governed by strict privacy regulations such as PHIPA~\cite{phipa2004} and GDPR~\cite{gdpr2016}. As a result, hospitals generally do not share raw patient records across institutions, limiting reproducibility and large-scale multi-hospital modeling. Synthetic EHR generation has emerged as a promising alternative for data augmentation and model training without exposing real patient records \cite{chen2025generating}. However, most existing synthetic EHR generators, such as MedGAN~\cite{choi2017generating}, TimeGAN~\cite{yoon2019time}, HALO~\cite{theodorou2023synthesize}, EHR-M-GAN~\cite{li2023generating}, FlexGen-EHR~\cite{flexgen}, TIMEDIFF~\cite{tian2024reliable}, and TarDiff~\cite{deng2025tardiff}, remain \emph{\textbf{centralized}} and rely on pooled multi-hospital data for training. This motivates \textit{privacy-preserving, cross-hospital synthetic generation of TS-EHRs without sharing raw patient data}.

Federated Learning (FL) enables privacy-preserving collaboration by aggregating local model updates instead of sharing raw patient data \cite{bai2025unified, fedweight}. Recent advances in federated generative modeling have enabled synthetic data generation without centralizing sensitive samples,  demonstrating utility in domains such as images, speech, and text \cite{hou2025private, xiong2023federated}. However, these methods are largely developed for single-modality, relatively homogeneous data and do not readily apply to clinical TS-EHRs, which are high dimensional, extreme sparse, and highly heterogeneous across hospitals. As a result, federated generative modeling for multi-hospital TS-EHR remains largely unexplored \cite{chen2025generating}, and to the best of our knowledge, no prior work has systematically addressed this setting.

Extending federated generative modeling to clinical TS-EHRs introduces unique challenges not present in centralized or conventional FL generative modeling.  First, raw TS-EHR features are inherently high-dimensional and extremely sparse \cite{theodorou2023synthesize}, often comprising thousands of binary indicators per timestamp after discretization into multi-hot representations  \cite{fiddle}. Such structural sparsity makes direct federated training of deep generative models unstable and prone to collapse, motivating dimensionality reduction prior to temporal generation modeling. Second, even after dimensionality reduction, hospitals exhibit substantial cross-hospital heterogeneity due to variations in patient populations, clinical workflows, coding practices, and measurement frequency \cite{EHR-Safe}. Such non-independent and identically distributed (non-IID) data, arising from covariate and temporal distribution shifts \cite{fedweight}, induce representation drift across hospitals, making standard federated aggregation strategies (e.g., FedAvg \cite{fedavg}) ineffective.

\begin{figure}[!t]
\centering
\setlength{\abovecaptionskip}{0.3cm}
\includegraphics[width=0.95\linewidth,scale=1.0]{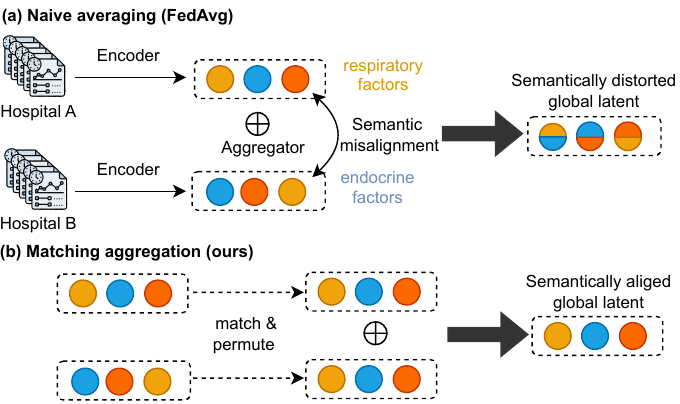}
\caption{Conceptual comparison of two aggregation strategies for federated latent encoders with respect to semantic alignment of the global latent space.
}
\label{fig:motivation}
\vspace{-0.4cm}
\end{figure}

Beyond data sparsity and distributional heterogeneity, effective time-series generative modeling further requires the latent space to be semantically aligned across hospitals \cite{theodorou2023synthesize}. 
Specifically, corresponding latent dimensions should encode consistent clinical semantics across hospitals, such that the same latent coordinate represents similar physiological or treatment-related factors rather than hospital-specific artifacts. Without such coordinate-wise alignment, naive federated aggregation produces a semantically distorted global latent space (Figure~\ref{fig:motivation}a). This misalignment hinders the learning of coherent temporal dependencies and ultimately degrades both the fidelity of synthetic trajectories and their downstream predictive utility. Together, these challenges highlight the fundamental limitations of standard FL methods, which are insufficient for multi-hospital TS-EHR generation task.

To address these limitations, we propose \textbf{FedEHR-Gen}, a federated framework for synthetic TS-EHR generation across distributed hospitals. The key idea of FedEHR-Gen is to decouple representation learning from temporal generative modeling, enabling robust and scalable generation under high-dimensional sparsity and cross-hospital heterogeneity. Specifically, FedEHR-Gen first learns a compact and semantically consistent latent representation across hospitals via matching aggregation, which mitigates training instability and representation drift induced by heterogeneous TS-EHR data (Figure~\ref{fig:motivation}b). 
Building upon this aligned latent space, the framework then performs federated temporal generative modeling in a distribution-aware manner to accommodate covariate and temporal shifts across hospitals.
This design enables stable synthetic TS-EHR generation without sharing raw patient data and is well suited for realistic multi-hospital federated settings.
We evaluate FedEHR-Gen on the eICU \cite{eicu} and MIMIC-III \cite{mimic3} datasets, covering diverse prediction tasks and heterogeneous hospital settings. 
These experiments assess the effectiveness of FedEHR-Gen in realistic federated healthcare scenarios.

Our main contributions are summarized as follows:
\ding{182} We propose \textbf{FedEHR-Gen}, a federated framework for synthetic TS-EHR generation that supports privacy-preserving multi-hospital collaboration.
\ding{183} We introduce a federated representation learning approach with \textbf{matching aggregation} to align encoding representations across hospitals, constructing a semantically consistent global latent space.
\ding{184} We develop a federated temporal generative model with \textbf{distribution-aware aggregation} to address cross-hospital data heterogeneity and stabilize temporal generative modeling.
\ding{185} Extensive experiments on eICU and MIMIC-III demonstrate that FedEHR-Gen achieves favorable generation fidelity, downstream predictive utility, and empirical privacy risk.

\begin{figure*}[!t]
\centering
\setlength{\abovecaptionskip}{0.1cm}
\includegraphics[width=0.85\linewidth]{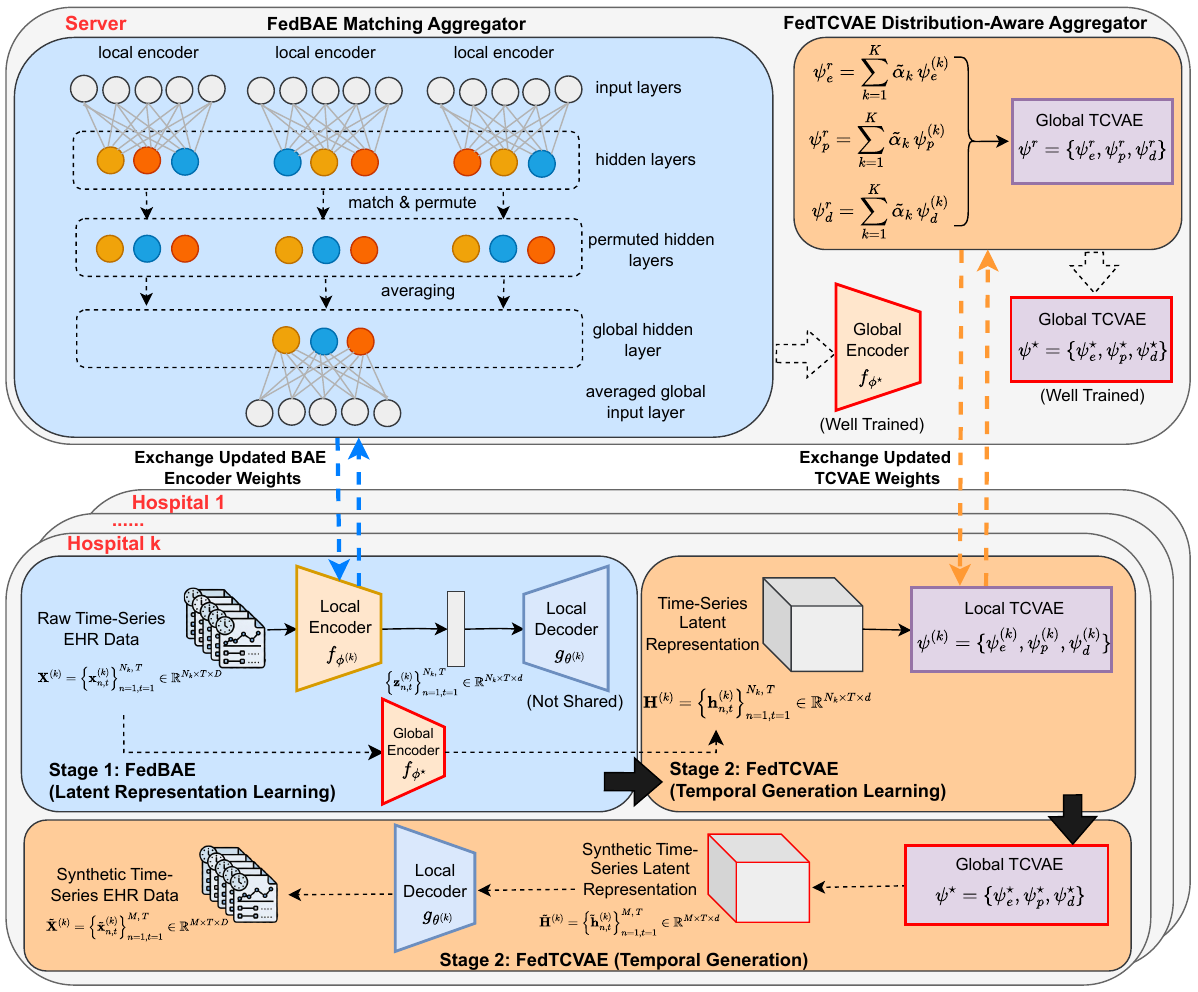}
\caption{Overview of the FedEHR-Gen framework. The framework follows a two-stage FL paradigm. In Stage 1 (FedBAE), hospitals train local BAEs on raw TS-EHR data and upload encoders to the server, where the matching aggregator averages them to obtain a global encoder for cross-hospital latent learning. In Stage 2 (FedTCVAE), hospitals train local TCVAE models on the aligned latent representations, and the server applies the distribution-aware aggregator to obtain a global TCVAE. The global TCVAE generates synthetic latent sequences, which are decoded by local decoders to reconstruct synthetic TS-EHR data. 
}
\label{fig:fedehr-gen}
\vspace{-0.3cm}
\end{figure*}

\section{Preliminaries} \label{sec:preliminaries}

We provide background on FL, Autoencoders (AEs) and Variational AEs (VAEs), and Temporal Conditional VAE (TCVAE) for time-series modeling. Additional related work is deferred to Appendix~\ref{appendix:related_work}.

\subsection{Federated Learning Setup and Notation}

We consider a cross-silo FL setting with $K$ hospitals, indexed by $k \in \{1,\dots,K\}$. 
Each hospital $k$ holds a private dataset of TS-EHR sequences
$\mathcal{D}^{(k)} = \{\mathbf{x}^{(k)}_{n,1:T_n}\}_{n=1}^{N_k}$,
where $\mathbf{x}^{(k)}_{n,t} \in \mathbb{R}^D$ denotes the feature vector of sample $n$ at time $t$, and $T_n$ is its sequence length.

A central server coordinates training by aggregating model updates from federated hospitals without accessing data.
We denote local model parameters at hospital $k$ by $\Theta^{(k)}$ and global parameters by $\Theta$. The objective of FL is to minimize the global empirical loss:
\begin{equation}
\min_{\Theta} \; \mathcal{L}(\Theta)
= \sum_{k=1}^{K} \frac{N_k}{N} \, \mathcal{L}^{(k)}(\Theta),
\end{equation}
where $\mathcal{L}^{(k)}(\Theta)$ is the local loss at hospital $k$, and $N = \sum_{k=1}^K N_k$.

Training proceeds in communication rounds. At each round, hospitals perform local updates
and transmit updated model parameters to the server, which computes a weighted aggregation:
\begin{equation}
\Theta \leftarrow \sum_{k=1}^{K} \frac{N_k}{N} \, \Theta^{(k)}.
\label{eq:fedavg}
\end{equation}

\subsection{Autoencoders and Variational Autoencoders}

AEs learn latent representations by coupling an encoder with a decoder \cite{bank2023autoencoders}.  Given an input $\mathbf{x}$, an encoder $f_\phi$ produces the corresponding  latent code $\mathbf{z}=f_\phi(\mathbf{x})$, 
and a decoder $g_\theta$ reconstructs the input as $\hat{\mathbf{x}}=g_\theta(\mathbf{z})$. 
AEs are trained by minimizing a reconstruction loss $\ell(\cdot,\cdot)$, such as mean squared error (MSE) for continuous data or binary cross-entropy (BCE) for binary features:
\begin{equation}
\mathcal{L}_{\mathrm{AE}}(\phi,\theta)
= \ell \bigl(\mathbf{x}, g_\theta(f_\phi(\mathbf{x})) \bigr).
\end{equation}

VAEs extend AEs into a probabilistic model by learning an approximate posterior 
$q_\phi(\mathbf{z}\mid \mathbf{x})$ and a generative decoder $p_\theta(\mathbf{x}\mid \mathbf{z})$, optimized via the evidence lower bound (ELBO) \cite{kingma2013auto}:
\begin{equation}
\mathcal{L}_{\mathrm{VAE}}(\mathbf{x};\phi,\theta)
= \mathbb{E}_{q_\phi(\mathbf{z}\mid\mathbf{x})}\bigl[\log p_\theta(\mathbf{x}\mid\mathbf{z})\bigr] - \mathrm{KL}\bigl(q_\phi(\mathbf{z}\mid\mathbf{x}) \,\|\, p(\mathbf{z})\bigr),
\end{equation}
where $p(\mathbf{z})$ is a prior distribution (e.g., Gaussian).

\subsection{Temporal Conditional VAEs} 

For time-series modeling, VAEs can be extended to sequential latent-variable models such as TCVAE \cite{he2024distributional,xu2023conditional}.
Given a time-series $\mathbf{x}_{1:T}$ and an optional condition $c$, TCVAE introduces a sequence of latent variables $\mathbf{z}_{1:T}$ and defines the joint distribution:
\begin{equation}
p(\mathbf{x}_{1:T}, \mathbf{z}_{1:T} \mid c)
= \prod_{t=1}^T p(\mathbf{z}_t \mid \mathbf{z}_{<t}, c)\,
p(\mathbf{x}_t \mid \mathbf{z}_t, \mathbf{z}_{<t}, c).
\end{equation}
The approximate posterior is factorized as
\begin{equation}
q(\mathbf{z}_{1:T} \mid \mathbf{x}_{1:T}, c)
= \prod_{t=1}^T q(\mathbf{z}_t \mid \mathbf{z}_{<t}, \mathbf{x}_{\le t}, c).
\end{equation}
The TCVAE is trained by maximizing the sequential ELBO over time. The generative and inference conditionals are  typically parameterized with recurrent or attention-based architectures.

\section{Methodology}  \label{sec:method}




This section introduces FedEHR-Gen and its main components.
FedEHR-Gen is a federated framework for synthetic TS-EHR generation under cross-hospital heterogeneity (Figure~\ref{fig:fedehr-gen}).
It adopts a two-stage paradigm that separates representation learning from temporal generation to stabilize training on high-dimensional and sparse EHRs. 
In the first stage, hospitals collaboratively learn a semantically aligned latent representation via federated binary AE (FedBAE) with a matching aggregation (\emph{MA}) approach.
In the second stage, hospitals train a federated temporal generator in this aligned latent space using federated TCVAE (FedTCVAE) with distribution-aware aggregation (\emph{DA}) to learn global temporal dynamics, while local decoding preserves hospital-specific characteristics.
The two stages are detailed in Sections~\ref{subsec:fedbae} and~\ref{subsec:fedtcvae}, respectively.

\subsection{Federated BAE} \label{subsec:fedbae}

Let hospital $k\in\{1,\dots,K\}$ hold preprocessed binary TS-EHR data $\mathbf{X}^{(k)}\in\{0,1\}^{N_k\times T\times D}$, where each observation $\mathbf{x}^{(k)}_{n,t}\in\{0,1\}^D$ is high-dimensional and sparse. Directly modeling temporal dynamics on such raw EHR data is often ill-posed and unstable,
especially under federated data partitioning across hospitals.
FedEHR-Gen therefore first learns a global BAE that maps local EHR data into a compact and semantically aligned latent space, which (i) reduces dimensionality and sparsity and (ii) ensures coordinate-wise semantic alignment across hospitals, making subsequent temporal modeling well-posed. This is achieved via a FedBAE with \emph{MA}. 

\paragraph{\textbf{Local BAE}}
Each hospital $k$ maintains a local encoder--decoder pair
$\bigl(f_{\phi^{(k)}}, g_{\theta^{(k)}}\bigr)$.
For an input $\mathbf{x}^{(k)}_{n,t}$, the encoder outputs a deterministic latent representation
\begin{equation}
\mathbf{z}^{(k)}_{n,t}
= f_{\phi^{(k)}}\!\left(\mathbf{x}^{(k)}_{n,t}\right)\in\mathbb{R}^{d},
\label{eq:ae_latent}
\end{equation}
and the decoder outputs Bernoulli parameters as
\begin{equation}
\hat{\mathbf{x}}^{(k)}_{n,t}
= g_{\theta^{(k)}}\!\left(\mathbf{z}^{(k)}_{n,t}\right)\in(0,1)^D .
\label{eq:ae_decoder_output}
\end{equation}
Hospital $k$ minimizes a binary reconstruction loss defined over its local dataset:
\begin{equation}
\mathcal{L}^{(k)}_{\mathrm{BAE}}\!\left(\phi^{(k)},\theta^{(k)}\right)
=
\frac{1}{N_k T}\sum_{n=1}^{N_k}\sum_{t=1}^{T}
\mathrm{BCE}\!\left(\mathbf{x}^{(k)}_{n,t},\hat{\mathbf{x}}^{(k)}_{n,t}\right).
\label{eq:ae_loss}
\end{equation}

In each federated round, hospitals learn locally well-optimized BAEs; however, due to independent optimization on heterogeneous local EHR data, the resulting latent representations may exhibit semantic misalignment across hospitals, often manifested as permutation-induced mismatches of latent coordinates.

\paragraph{\textbf{Why does FedAvg fail to aggregate encoders?}}
The semantic misalignment observed across hospitals arises from the permutation invariance of neural hidden units \cite{entezari2022the}. For example, two encoders can represent functionally similar mappings while assigning different semantic meanings to latent coordinates.
Consider the weight matrices of a given hidden layer (or the latent layer) from two hospitals, denoted by $\mathbf{W}^{(k)}$ and $\mathbf{W}^{(j)}$.
Under heterogeneous local data, their learned neuron orderings may differ by permutation matrices $\mathbf{\Pi}_k \neq \mathbf{\Pi}_j$, such that
$\mathbf{W}^{(k)} = \mathbf{W}\mathbf{\Pi}_k$ and $\mathbf{W}^{(j)} = \mathbf{W}\mathbf{\Pi}_j$.
Coordinate-wise averaging as performed by FedAvg then yields
\begin{equation}
\bar{\mathbf{W}}
= \frac{1}{2}\left(\mathbf{W}^{(k)} + \mathbf{W}^{(j)}\right)
= \frac{1}{2}\left(\mathbf{W}\mathbf{\Pi}_k + \mathbf{W}\mathbf{\Pi}_j\right),
\label{eq:fedavg_failure} 
\end{equation}
which, with high probability, does \emph{not} correspond to any valid permutation of a coherent encoder, i.e.,
$\bar{\mathbf{W}} \neq \mathbf{W}\mathbf{\Pi}$ for all permutation matrices $\mathbf{\Pi}$ (Figure \ref{fig:motivation}a).
If the permutations were known, one could first realign the encoders before averaging:
\begin{equation}
\frac{1}{2}\left(\mathbf{W}^{(k)}\mathbf{\Pi}_k^\top + \mathbf{W}^{(j)}\mathbf{\Pi}_j^\top\right)
= \mathbf{W},
\label{eq:undo_permutation}
\end{equation}
thereby recovering a semantically coherent encoder and motivating explicitly estimating layer-wise permutations prior to aggregation.

\paragraph{\textbf{Matching aggregation for encoder alignment.}}

To address this issue, we adopt a layer-wise \emph{MA} strategy inspired by \cite{fedma}, which explicitly aligns encoder hidden neurons prior to averaging (Figure~\ref{fig:motivation}b). Consider an encoder hidden layer $\ell \in \{1,\dots,L\}$. Let $\mathbf{W}^{(k)}_{\ell}\in\mathbb{R}^{m_{\ell-1}\times m_{\ell}}$ denote the weight matrix of layer $\ell$ at hospital $k$, where $m_{\ell-1}$ and $m_{\ell}$ are the input and output widths of this layer, respectively.
We view each hidden neuron as a column vector.
Specifically, the $i$-th neuron learned at hospital $k$ is
\begin{equation}
\mathbf{w}^{(k)}_{\ell,i} = \mathbf{W}^{(k)}_{\ell}(:, i)\in\mathbb{R}^{m_{\ell-1}}.
\label{eq:nueron_vector}
\end{equation}
Let $\tilde{\mathbf{w}}_{\ell,j}$ denote the $j$-th neuron of a reference encoder that defines a canonical ordering (e.g., the global encoder from the previous communication round). Given a neuron similarity cost $c(\cdot,\cdot)$, the server solves a bipartite matching problem to align hospital $k$'s neurons to the reference:
\begin{equation}
\min_{\{\pi^{(k)}_{i,j}\}}
\sum_{i}\sum_{j}
\pi^{(k)}_{i,j}\,
c\!\left(\mathbf{w}^{(k)}_{\ell,i}, \tilde{\mathbf{w}}_{\ell,j}\right),
\quad
\text{s.t.}\;
\sum_{j}\pi^{(k)}_{i,j}=1,\;
\sum_{i}\pi^{(k)}_{i,j}=1,
\label{eq:matching_obj}
\end{equation}
where $\pi^{(k)}_{i,j}\in\{0,1\}$ induces a permutation matrix $\mathbf{\Pi}^{(k)}_{\ell}\in\{0,1\}^{m_{\ell}\times m_{\ell}}$ that reorders the output units of layer $\ell$ at hospital $k$ to match the reference ordering, which is the federated average at the previous iteration. 
We solve Eq.~\eqref{eq:matching_obj} efficiently using the Hungarian algorithm \cite{kuhn1955hungarian}. To start the algorithm, we set the reference to be the standard federated average. Alternatively, we can choose an anchor hospital as the reference and align the rest of the hospitals against it. In our main results, we opted for the former approach, confirming that the two approaches produce similar results. Further details on the initial reference selection are provided in Appendix~\ref{appendix:ref-anchor}.

After alignment, the aggregated encoder layer is obtained by permutation-aware averaging:
\begin{equation}
\bar{\mathbf{W}}_{\ell}
=
\sum_{k=1}^{K}
\alpha_k\,
\mathbf{W}^{(k)}_{\ell}\mathbf{\Pi}^{(k)}_{\ell}, 
\label{eq:matched_avg}
\end{equation}
where  $\alpha_k = \frac{N_k}{\sum_{k'=1}^{K}N_{k'}}$. To preserve functional consistency, the same permutation is applied to the input channels of the next layer.

Applying this aggregation to all encoder hidden layers, including the final latent layer, yields the global encoder
$f_{\phi^\star}=\{ \bar{\mathbf{W}}_{\ell} \}_{\ell=1}^{L}$.

\paragraph{\textbf{Local decoder adaptation.}}
After server-side \emph{MA}, the server broadcasts the aligned global encoder $\phi^\star$ to all hospitals, and each hospital replaces its local encoder accordingly, $\phi^{(k)} \leftarrow \phi^\star$. 
To maintain consistency with the permuted latent coordinates produced by the aligned global encoder, we apply the corresponding permutation to the latent-to-hidden mapping of the decoder, thereby maintaining the consistency of the encoder-decoder interface. The encoder is then frozen, and only the local decoder is fine-tuned to preserve hospital-specific reconstruction:
\begin{equation}
\min_{\theta^{(k)}} \;\; \mathcal{L}^{(k)}_{\mathrm{BAE}}\!\left(\phi^\star,\theta^{(k)}\right).
\label{eq:decoder_fine_tuning}
\end{equation}

\paragraph{\textbf{Aligned latent representations.}}
After encoder alignment, each hospital constructs its latent representation by
forwarding local EHR sequences through the global encoder:
\begin{equation}
\mathbf{H}^{(k)}
=
f_{\phi^\star}\!\left(\mathbf{X}^{(k)}\right)
=
\left\{
\mathbf{h}^{(k)}_{n,t}
\right\}_{n=1,t=1}^{N_k,\,T}
\in \mathbb{R}^{N_k \times T \times d},
\label{eq:latent_tensor}
\end{equation}
where $\mathbf{h}^{(k)}_{n,t}\in\mathbb{R}^d$ denotes the aligned  latent embedding at time $t$. These aligned latent representations provide a shared input space for subsequent
federated temporal generative modeling.
The detailed implementation is provided in Appendix~\ref{appendix:implementation} \textbf{Algorithm}~\ref{alg:fedbae}.

\subsection{Federated TCVAE} \label{subsec:fedtcvae}

Building upon the semantically aligned latent representations, we perform federated temporal generative modeling directly in the latent space. Let $\mathbf{H}^{(k)}$ denote the aligned latent tensor at hospital $k$, with $\mathbf{h}^{(k)}_{n,1:T}$ representing the latent trajectory of sample $n$.
Compared to raw EHR observations, these latent sequences are compact, continuous, and semantically consistent across hospitals, enabling stable federated temporal modeling.

\paragraph{\textbf{Local TCVAE}}
To capture temporal dependencies in the learned representations,
we adopt a TCVAE with recurrent state transitions. At each time step $t$, a recurrent hidden state $\mathbf{s}_t$
summarizes the historical information up to time $t-1$:
\begin{equation}
\mathbf{s}_t
=
\mathrm{RNN}\!\left(\mathbf{s}_{t-1}, \mathbf{h}_{t-1}, \mathbf{c}\right),
\label{eq:tcvae_rnn}
\end{equation}
where $\mathbf{h}_{t-1}$ denotes the latent representation obtained
from the BAE at the previous time step, and $\mathbf{c}$ represents conditioning variables shared across time.

Conditioned on $\mathbf{s}_t$ and $\mathbf{c}$, the temporal prior, approximate posterior (encoder) network, and generative likelihood (decoder) network are defined as 
\begin{equation}
\left\{
\begin{aligned}
\text{Prior:}\quad & p_{\psi_p}\!\left(z_t \mid s_t, c\right),\\
\text{Posterior:}\quad & q_{\psi_e}\!\left(z_t \mid h_t, s_t, c\right),\\
\text{Likelihood:}\quad & p_{\psi_d}\!\left(h_t \mid z_t, s_t, c\right).
\end{aligned}
\right.
\label{eq:model_components}
\end{equation}
where $\psi=\{\psi_e,\psi_p,\psi_d\}$ denotes the set of parameters of the
encoder, temporal prior, and decoder, respectively.

At hospital $k$, TCVAE is trained by minimizing the ELBO:
\begin{equation}
\resizebox{0.95\linewidth}{!}{$
\begin{aligned}
\mathcal{L}^{(k)}_{\mathrm{TCVAE}}
&=
\frac{1}{N_k}
\sum_{n=1}^{N_k}
\sum_{t=1}^{T}
\Bigl[
-
\mathbb{E}_{q_{\psi_e}\!\left(
\mathbf{z}^{(k)}_{n,t} \mid
\mathbf{h}^{(k)}_{n,t},
\mathbf{s}^{(k)}_{n,t},
\mathbf{c}^{(k)}_n
\right)}
\log p_{\psi_d}\!\left(
\mathbf{h}^{(k)}_{n,t} \mid
\mathbf{z}^{(k)}_{n,t},
\mathbf{s}^{(k)}_{n,t},
\mathbf{c}^{(k)}_n
\right)
\\
&\quad
+
\lambda\,
\mathrm{KL}\!\left(
q_{\psi_e}\!\left(
\mathbf{z}^{(k)}_{n,t} \mid
\mathbf{h}^{(k)}_{n,t},
\mathbf{s}^{(k)}_{n,t},
\mathbf{c}^{(k)}_n
\right)
\,\bigg\|\,
p_{\psi_p}\!\left(
\mathbf{z}_t \mid
\mathbf{s}^{(k)}_{n,t},
\mathbf{c}^{(k)}_n
\right)
\right)
\Bigr],
\end{aligned}
\label{eq:local_tcvae_loss}
$}
\end{equation}
where $\lambda$ controls posterior regularization. We provide the overall training and generation procedures of the TCVAE in Appendix~\ref{appendix:implementation}, including Figure~\ref{fig:tcvae_train_gen}, Algorithm~\ref{alg:local_tcvae_train}, and Algorithm~\ref{alg:global_tcvae_gen}.


\paragraph{\textbf{Distribution-aware aggregation.}}
Optimizing the local TCVAE induces a hospital-specific temporal latent distribution,
\begin{equation}
q^{(k)}_t(\mathbf{z})
=
\frac{1}{N_k}
\sum_{n=1}^{N_k}
q_{\psi_e}\!\left(
\mathbf{z} \mid \mathbf{h}^{(k)}_{n,\le t},
\mathbf{s}^{(k)}_{n,t}, \mathbf{c}^{(k)}_n
\right),
\label{eq:latent_dist}
\end{equation}
which captures hospital-dependent temporal trajectories.
Direct parameter averaging is suboptimal under cross-hospital distribution shifts, as it ignores heterogeneity in temporal latent distributions. We therefore introduce \emph{DA} based on latent distribution similarity.
For hospitals $k$ and $j$, we define the temporal latent divergence as
\begin{equation}
d_{k,j}
=
\frac{1}{T}
\sum_{t=1}^{T}
\mathrm{KL}\!\left(
q^{(k)}_t(\mathbf{z}) \,\|\, q^{(j)}_t(\mathbf{z})
\right),
\label{eq:latent_div}
\end{equation}
where the KL divergence reflects how well the latent distribution
of hospital $j$ can approximate that of hospital $k$.
Based on the divergence of hospital $j$ with all hospitals, we compute \emph{DA} weights:
\begin{equation}
\tilde{\alpha}_k
=
\frac{
\alpha_k \exp\!\left(-\tau \bar{d}_k\right)
}{
\sum_{j=1}^{K} \alpha_j \exp\!\left(-\tau \bar{d}_j\right)
},
\label{eq:dist_weight}
\end{equation}
where $\tau > 0$ controls the sensitivity to distribution mismatch, and $\bar{d}_k = \frac{1}{K-1} \sum_{j \neq k} d_{k,j}$ denotes the average latent distribution divergence of hospital $k$ to the others. 

Using these weights, we perform component-wise aggregation:
\begin{equation}
\psi_\bullet^r = \sum_{k=1}^{K} \tilde{\alpha_k} \, \psi_\bullet^{(k)}, \;
\bullet \in \{e,p,d\},
\label{eq:tcvae_agg}
\end{equation}
where $r$ denotes the index of the federated communication round.
This aggregation reflects that all components are aligned and weighted according to cross-hospital latent distribution similarity.
The resulting global temporal model is
$\psi^r = \{\psi_e^r, \psi_p^r, \psi_d^r\}$,
which is iteratively updated across communication rounds. 

The above proposed approach down-weights out-of-distribution hospitals relative to the majority hospitals in order to maintain strong overall performance across all sites. In future work, we will adopt a hospital-specific aggregation strategy, in which each hospital benefits by borrowing appropriately weighted information from other hospitals \cite{fedweight}. Specifically, each hospital receives the aligned parameters (Eq. \ref{eq:matching_obj}) from all participating hospitals. Rather than performing federated averaging on a central server, distribution-aware federated averaging is carried out locally within each hospital.

\paragraph{\textbf{Synthetic EHR generation via local decoding.}}
After convergence, the global TCVAE $\psi^\star = \{\psi_e^\star, \psi_p^\star, \psi_d^\star\}$ defines a generator over latent trajectories.
At each hospital, temporal latent variables
$\{\tilde{\mathbf{z}}_t\}_{t=1}^{T}$ are recursively sampled from the learned temporal prior
$p_{\psi_p^\star}(\mathbf{z}_t \mid \mathbf{z}_{<t},c)$,
and aligned latent representations
$\tilde{\mathbf{h}}_{1:T}$ are generated according to
Eqs.~\eqref{eq:tcvae_rnn}--\eqref{eq:model_components}.
Each hospital then applies its local decoder
$g_{\theta^{(k)}}$ to map the generated
latent sequences back to the observation space, yielding
hospital-specific synthetic EHR:
\begin{equation}
\tilde{\mathbf{x}}^{(k)}_{1:T}
=
g_{\theta^{(k)}}\!\left(\tilde{\mathbf{h}}_{1:T}\right).
\label{eq:local_decode}
\end{equation}
We detail its implementation in Appendix~\ref{appendix:implementation} \textbf{Algorithm}~\ref{alg:fedtcvae}.

\section{Experiments}
\label{sec:experiments}

\paragraph{\textbf{Datasets}}
We evaluate FedEHR-Gen on two publicly available real-world EHR datasets: MIMIC-III \cite{mimic3} and eICU \cite{eicu}. Using the FIDDLE preprocessing pipeline~\cite{fiddle}, we obtain two clinical risk prediction cohorts, namely ARF (acute respiratory failure)-4H and Mortality-48H. Cohort sizes and feature dimensions for each task are summarized in Appendix
Table~\ref{tab:dataset_statistics}.
For eICU, samples are grouped by their source hospitals to naturally form federated clients in a cross-silo FL setting (Appendix~\ref{appendix:datasets}; Figure~\ref{fig:sample-distribution-over-hips}). 
Additional dataset and preprocessing details are provided in Appendix~\ref{appendix:datasets}.

\paragraph{\textbf{FL Protocol}}
We adopt a standard cross-silo FL protocol with a central server coordinating model updates. Training proceeds in communication rounds, where all hospitals participate in each round.
Hyperparameters such as the number of communication rounds, local epochs, and batch size are fixed across methods for fair comparison. 
Implementation details are provided in Appendix~\ref{appendix:implementation}.

\paragraph{\textbf{Baselines}}

We compare FedEHR-Gen with two representative baselines: (1) 
A centralized temporal generative model trained on pooled data serves as an upper-bound reference, denoted as \emph{Centralized}; (2) A federated baseline trains the same model using standard FedAvg, denoted as \emph{FedAvg}. These baselines enable a direct evaluation of the benefits of FedEHR-Gen.

\paragraph{\textbf{Evaluation Metrics}}

Following prior work \cite{choi2017generating, theodorou2023synthesize, li2023generating, tian2024reliable}, we evaluate the proposed method from three complementary perspectives: generation fidelity, downstream predictive utility, and privacy risk. Formal definitions of all metrics are provided in
Appendix~\ref{appendix:metrics}.

\textit{Generation Fidelity}
We assess the statistical and temporal fidelity of synthetic EHR data by comparing real and synthetic sequences in terms of marginal feature distributions and temporal correlations. Distributional similarity is quantified using Coefficient of Determination ($R^2$) (higher is better) and maximum mean discrepancy (MMD; lower is better). 

\textit{Downstream Predictive Utility}
To evaluate practical utility, we train downstream clinical prediction models using real, synthetic, and hybrid datasets. 
Model performance is evaluated on held-out real global test sets using Area Under the Precision--Recall Curve (AUPRC) due to the severe class imbalance, together with Area Under the Receiver Operating Characteristic Curve (AUROC).

\textit{Privacy Risk Assessment}
Following \cite{flexgen, tian2024reliable}, we assess privacy risk using attack-based empirical metrics that quantify the potential leakage of sensitive information from synthetic data. We report Membership Inference Risk (MIR) \cite{liu2019socinf} and the Nearest-Neighbor Adversarial Accuracy Risk (NNAA)  \cite{yale2020generation}. Lower MIR and NNAA indicate reduced privacy leakage, as they reflect how easily an attacker can infer the presence of individual training records or find synthetic samples that closely resemble real ones.

\section{Results}
\label{sec:main_results}

\subsection{Fidelity of the Generated Data}  
Generation fidelity is assessed from both feature-level consistency and distributional similarity using $R^2$ and MMD, respectively (\autoref{tab:fidality-c5-r2-mmd}). 
Centralized training provides an upper bound, achieving the highest $R^2$ and lowest MMD across both tasks. 
Under federated settings, FedEHR-Gen consistently improves generation fidelity over FedAvg. For ARF-4H, FedEHR-Gen increases $R^2$ from $0.672$ to $0.721$ and improves MMD from $0.892$ to $0.863$. For Mortality-48H, similar gains are observed, with $R^2$ improving from $0.651$ to $0.701$ and MMD decreasing from $0.921$ to $0.848$. 

\begin{table}[!t]
\centering
\setlength{\abovecaptionskip}{4pt}
\setlength{\belowcaptionskip}{0pt}
\fontsize{8}{10}\selectfont
\caption{Generation fidelity measured by $R^2$ ($\uparrow$) and MMD ($\downarrow$) on five eICU hospitals.}
\label{tab:fidality-c5-r2-mmd}
\setlength{\tabcolsep}{1pt}
\renewcommand{\arraystretch}{1.0}
\begin{tabular}{lcccc}
\toprule
\multirow{2}{*}{Dataset} 
& \multicolumn{2}{c}{ARF-4H} 
& \multicolumn{2}{c}{Mortality-48H}  \\
\cmidrule(lr){2-3} \cmidrule(lr){4-5} 
& $R^2$ ($\uparrow$) & MMD ($\downarrow$)
& $R^2$ ($\uparrow$) & MMD ($\downarrow$) \\
\midrule
Centralized 
& $0.768 \pm 0.038$ & $0.842 \pm 0.020$
& $0.712 \pm 0.029$ & $0.886 \pm 0.028$ \\

FedAvg 
& $0.672 \pm 0.041$ & $0.892 \pm 0.021$
& $0.651 \pm 0.043$ & $0.921 \pm 0.039$ \\

\textbf{FedEHR-Gen}
& $\mathbf{0.721 \pm 0.040}$ & $\mathbf{0.863 \pm 0.024}$
& $\mathbf{0.701 \pm 0.031}$ & $\mathbf{0.848 \pm 0.030}$ \\
\bottomrule
\end{tabular}
\vspace{-0.3cm}
\end{table}

\begin{figure}[!t]
\centering
\setlength{\abovecaptionskip}{0.1cm}
\includegraphics[width=\linewidth,scale=1.0]{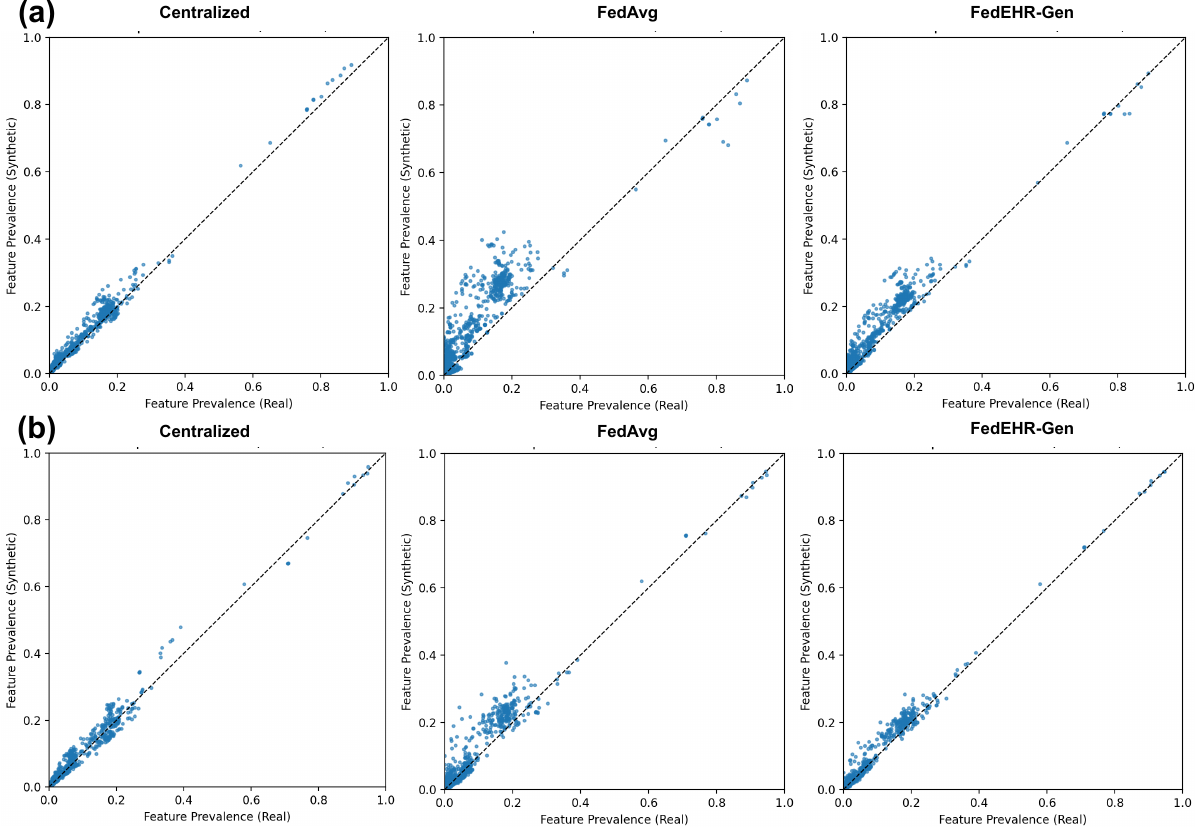}
\caption{Generation fidelity of averaged feature-wise prevalence on five eICU hospitals. (a) ARF-4H; (b) Mortality-48H.}
\label{fig:fidelity-feature-wise-statis-c5}
\vspace{-0.3cm}
\end{figure}

In addition, we examine the feature-wise prevalence of the generated data (\autoref{fig:fidelity-feature-wise-statis-c5}). As expected, centralized training exhibits the closest alignment, indicating the most faithful preservation of feature-wise statistics. In contrast, FedAvg shows increased dispersion around the diagonal, particularly for low-prevalence features. FedEHR-Gen consistently reduces this deviation across both ARF-4H and Mortality-48H, reflecting improved preservation of feature-wise prevalence. We also perform a per-timestamp feature-wise prevalence comparison of ARF-4H and observed similar results (Appendix~\ref{appendix:gen-fidelity}; Figure~\ref{fig:fidelity-arf4h-per-time-feature-prev}).

\begin{figure}[!t]
\centering
\setlength{\abovecaptionskip}{0.1cm}
\includegraphics[width=\linewidth,scale=1.0]{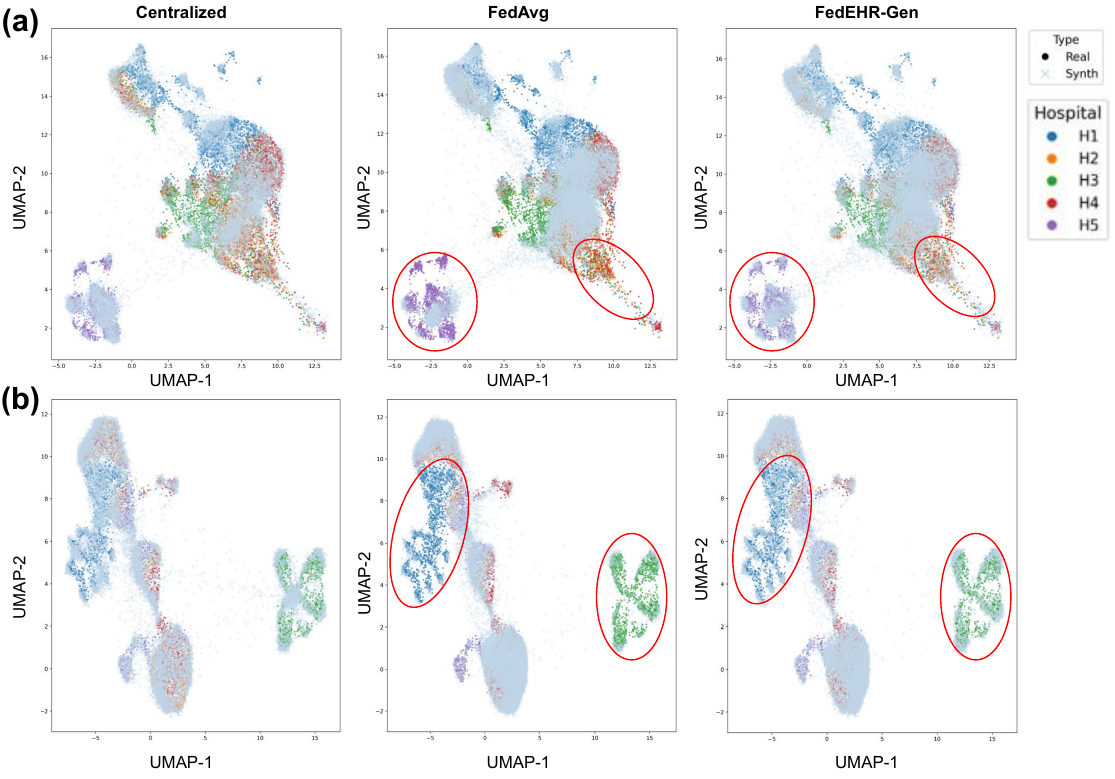}
\caption{Generation fidelity of UMAP visualization on five eICU hospitals. (a) ARF-4H; (b) Mortality-48H.}
\label{fig:fidelity-c5-umap}
\vspace{-0.3cm}
\end{figure}

We further examine the generated data using UMAP visualizations (\autoref{fig:fidelity-c5-umap}). Centralized training shows strong overlap between synthetic and real samples across hospitals. Under FedAvg, synthetic samples from certain hospitals are sparsely represented or partially missing in the embedding space (\autoref{fig:fidelity-c5-umap} circled regions). In contrast, FedEHR-Gen better recovers these hospital-specific patterns for both ARF-4H and Mortality-48H, indicating improved coverage of hospital-specific representation patterns observed in real data. Similarly, the corresponding per-timestamp UMAP visualizations exhibit consistent patterns (Appendix~\ref{appendix:gen-fidelity}; Figure~\ref{fig:fidelity-arf4h-per-time-umap}).

\begin{figure}[!t]
\centering
\setlength{\abovecaptionskip}{0.1cm}
\includegraphics[width=\linewidth,scale=1.0]{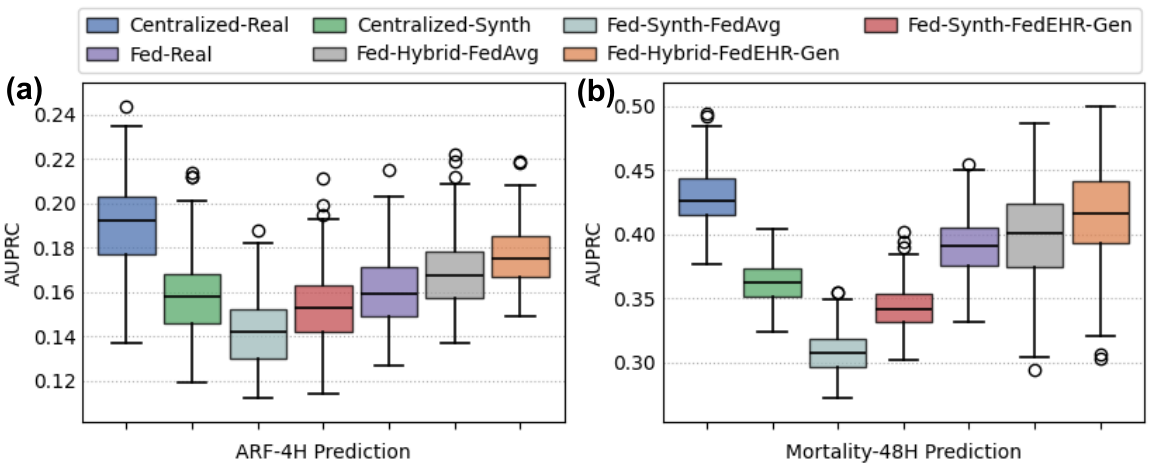}
\caption{Generation utility on five eICU hospitals. AUPRC for (a) ARF-4H prediction; (b) Mortality-48H prediction.}
\label{fig:fig:utility_synth_hybrid_c5_auprc}
\vspace{-0.3cm}
\end{figure}

\subsection{Clinical Tasks using the Generated Data} 

We evaluate 7 training strategies under both centralized and federated settings: 
(1) \textit{Centralized-Real} trains the prediction model on real data pooled from all hospitals in a centralized manner and serves as an upper bound;
(2) \textit{Centralized-Synth} trains the prediction model solely on synthetic data generated by a centrally trained generative model;
(3) \textit{Fed-Real} trains the prediction model on real data using standard FL across hospitals;
For synthetic-only federated baselines, (4) \textit{Fed-Synth-FedAvg} trains the generative model using FedAvg and then trains the prediction model on the resulting synthetic data, while (5) \textit{Fed-Synth-FedEHR-Gen} replaces FedAvg with our proposed federated generative framework;
For hybrid strategies, (6) \textit{Fed-Hybrid-FedAvg} augments each hospital’s local real dataset with an equal amount of synthetic data generated by a FedAvg-trained model, and (7) \textit{Fed-Hybrid-FedEHR-Gen} follows the same protocol using FedEHR-Gen.
Unless otherwise specified, all prediction models are evaluated on a global real test set.

As expected, models trained on real centralized data achieve the highest AUPRC across both tasks, serving as an upper bound for predictive performance (\autoref{fig:fig:utility_synth_hybrid_c5_auprc}). Training solely on synthetic data results in a consistent performance drop.  Importantly, hybrid training under federated settings improves prediction performance over both Fed-Real and Fed-Synth strategies, indicating that synthetic data is complementary to the real data. Notably, Fed-Hybrid-FedEHR-Gen attains the best performance among them, improving AUPRC from approximately 0.15 to 0.17 on ARF-4H and from 0.34 to 0.41 on Mortality-48H compared with its FedAvg-based counterpart, highlighting the utility gain brought by higher-fidelity synthetic data (\autoref{fig:fig:utility_synth_hybrid_c5_auprc}). Consistent performance trends are also observed when evaluated using AUROC (Appendix~\ref{appendix:utility}; Figure~\ref{fig:fig:utility_synth_hybrid_c5_auroc}).

\begin{figure}[!t]
\centering
\setlength{\abovecaptionskip}{0.1cm}
\includegraphics[width=\linewidth,scale=1.0]{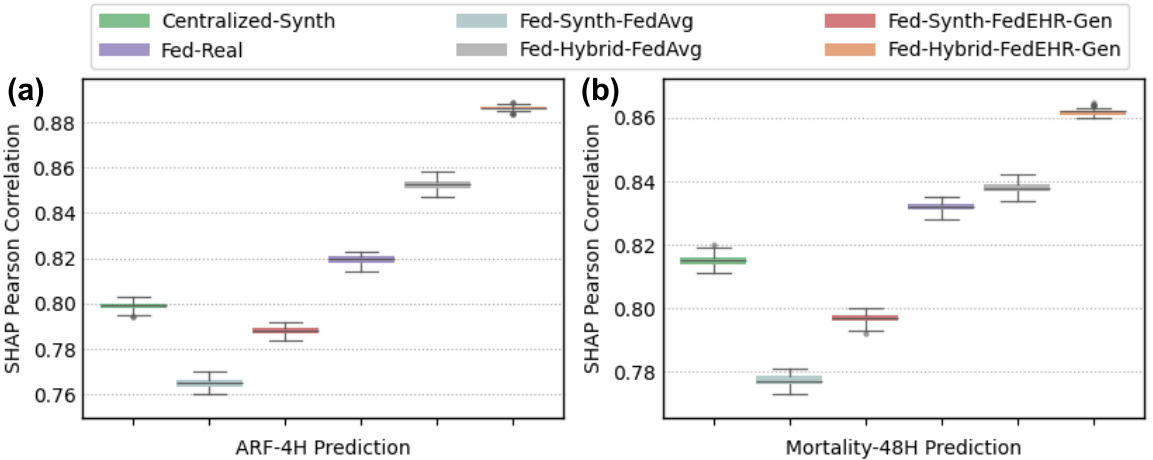}
\caption{Generation utility of Pearson correlation of SHAP Values with Centralized-Real on five eICU hospitals. (a) ARF-4H prediction; (b) Mortality-48H prediction.}
\label{fig:fig:utility_synth_hybrid_c5_shap}
\vspace{-0.3cm}
\end{figure}

From an explainable AI perspective, we further assess whether the generated data preserve feature importance implicated in the real data. To that end, we compute the Pearson correlation of SHapley Additive exPlanations (SHAP) values \cite{shapvalue} with those obtained under centralized training (\autoref{fig:fig:utility_synth_hybrid_c5_shap}). Models trained solely on synthetic data exhibit reduced SHAP agreement, with correlations around 0.76--0.79, indicating deviations in learned feature importance. Hybrid training consistently improves SHAP alignment, and Fed-Hybrid-FedEHR-Gen achieves the highest correlation among federated approaches, increasing SHAP correlation to approximately 0.85 for ARF-4H and 0.86 for Mortality-48H. These results suggest that higher-fidelity synthetic data not only improve predictive performance but also identify more important features in cross-hospital learning (\autoref{fig:fig:utility_synth_hybrid_c5_shap}).

\subsection{Privacy Preservation} 
We assess privacy preservation of the generated data using MIR and NNAA, where lower values indicate less privacy leakage between synthetic and real data (\autoref{tab:generation_privacy}). Centralized training yields the lowest leakage risk on both tasks. Under federated generation, FedAvg exhibits higher MIR and NNAA, indicating greater privacy exposure compared to centralized baselines. Compared with FedAvg, FedEHR-Gen consistently lowers privacy risk: on ARF-4H, MIR drops from 0.262 to 0.236 (approximately 10\%) and NNAA from 0.011 to 0.008 (approximately 27\%); on Mortality-48H, MIR decreases from 0.317 to 0.251 (approximately 21\%) and NNAA from 0.013 to 0.009 (approximately 31\%). Overall, FedEHR-Gen exhibits lower empirical privacy leakage than the baselines.

\begin{figure*}[!t]
\centering
\setlength{\abovecaptionskip}{0.1cm}
\includegraphics[width=0.9\linewidth,scale=1.0]{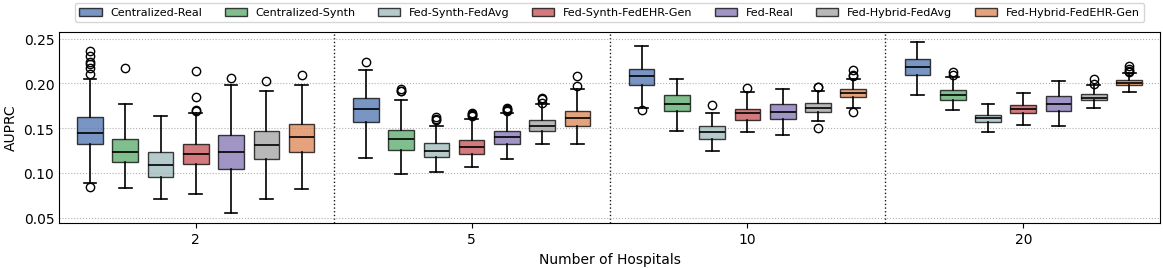}
\caption{Effect of varying federated scales (number of hospitals) on ARF-4H AUPRC on eICU across seven methods.
}
\label{fig:c5-scale-auprc}
\vspace{-0.3cm}
\end{figure*}

\subsection{Scalability Analysis}
\label{subsec:scalability}

We analyze the scalability of federated generation and downstream prediction under varying federated network sizes. Specifically, we consider four federated configurations with 2, 5, 10, and 20  eICU training hospitals. To ensure a fair and consistent evaluation across different scales, all models are evaluated on a unified real test set constructed by aggregating held-out data from the eICU hospitals 21--30. This protocol decouples changes in test distribution from the federated scale and allows performance differences to be attributed solely to the number of training hospitals.



As the federated scale increases from 2 to 20 hospitals, downstream AUPRC on the ARF-4H task exhibits a consistent upward trend across all methods (Figure~\ref{fig:c5-scale-auprc}).  For federated hybrid approaches, AUPRC increases from around 0.14--0.15 in the two-hospital setting to approximately 0.20--0.22 when 20 hospitals participate, indicating steady performance gains as more hospital data becomes available.  Similar monotonic improvements are observed for real-only and synthetic-only federated baselines, suggesting that expanding the number of hospitals leads to a richer joint data distribution that benefits downstream prediction. 
Across all scales, hybrid training strategies consistently outperform their real-only and synthetic-only counterparts, with \textit{Fed-Hybrid-FedEHR-Gen} achieving the highest AUPRC among federated approaches. 
Consistent performance trends are also validated on the Mortality-48H task (Appendix~\ref{appendix:scalability}; Figure~\ref{fig:c5-scale-auprc-mortality}).
These results indicate that high-fidelity synthetic data can support scalable performance improvements in large cross-hospital FL systems.

\begin{table}[!t]
\setlength{\abovecaptionskip}{4pt}
\setlength{\belowcaptionskip}{0pt}
\centering
\fontsize{8}{10}\selectfont
\caption{\textbf{Generation Privacy}. MIR Score ($\downarrow$) and NNAA Score ($\downarrow$) between synthetic and real datasets on five eICU hospitals.}
\label{tab:generation_privacy}
\setlength{\tabcolsep}{1.5pt}
\renewcommand{\arraystretch}{1.0}
\begin{tabular}{lcccc}
\toprule
\multirow{2}{*}{Dataset} 
& \multicolumn{2}{c}{ARF-4H} 
& \multicolumn{2}{c}{Mortality-48H}  \\
\cmidrule(lr){2-3} \cmidrule(lr){4-5} 
& MIR ($\downarrow$) & NNAA ($\downarrow$)
& MIR ($\downarrow$) & NNAA ($\downarrow$) \\
\midrule
Centralized 
& $0.214 \pm 0.018$ & $0.006 \pm 0.001$
& $0.198 \pm 0.016$ & $0.005 \pm 0.001$ \\

FedAvg 
& $0.262 \pm 0.034$ & $0.011 \pm 0.003$
& $0.317 \pm 0.031$ & $0.013 \pm 0.002$ \\

\textbf{FedEHR-Gen}
& $\mathbf{0.236 \pm 0.022}$ & $\mathbf{0.008 \pm 0.001}$
& $\mathbf{0.251 \pm 0.021}$ & $\mathbf{0.009 \pm 0.002}$ \\
\bottomrule
\end{tabular}
\vspace{-0.6cm}
\end{table}

\subsection{Cross-Dataset Federated Generation}
\label{subsec:cross_dataset}
\begin{figure}[!t]
\centering
\setlength{\abovecaptionskip}{0.1cm}
\includegraphics[width=0.95\linewidth,scale=1.0]{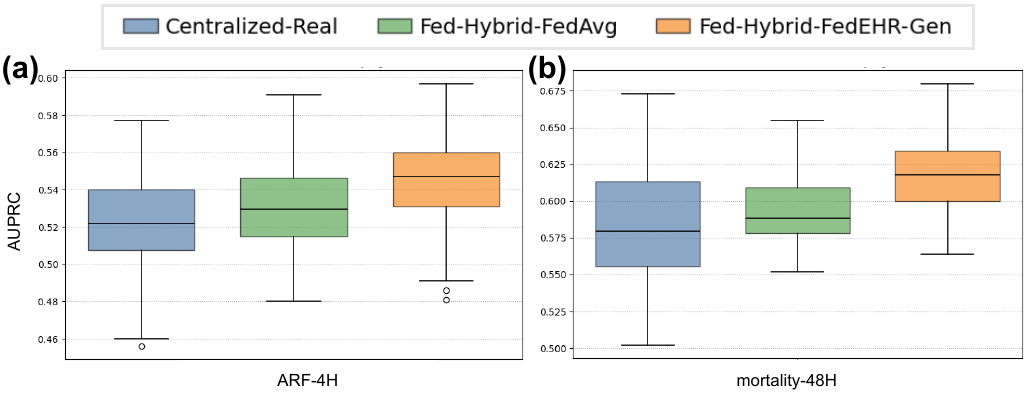}
\caption{Cross-dataset generation utility on MIMIC-III. AUPRC for models trained with synthetic data generated by a generator trained on eICU and evaluated on MIMIC-III.
(a) ARF-4H prediction; (b) Mortality-48H prediction.}

\label{fig:cross-datasets-c5}
\vspace{-0.3cm}
\end{figure}

We explore federated training across heterogeneous datasets by treating eICU and MIMIC-III as distinct clients, and evaluating cross-dataset
generalization under domain shift. Specifically, we train a FedTCVAE model on eICU to learn a global latent temporal generator, which is then used to generate synthetic latent sequences that are decoded by the local MIMIC-III decoder. The resulting synthetic data are combined with real MIMIC-III data for downstream model training. Consistent with observations on the eICU dataset, hybrid training on MIMIC-III yields improved generation utility across both ARF-4H and Mortality-48H tasks (\autoref{fig:cross-datasets-c5}). In particular,
Fed-Hybrid-FedEHR-Gen achieves higher AUPRC than real-only federated baselines, indicating that synthetic data provide complementary predictive signals. 
A consistent trend is also observed when evaluated using AUROC (Appendix~\ref{appendix:cross-dataset}; Figure~\ref{fig:cross-datasets-c5-auroc}).
These results suggest that the proposed generation strategy generalizes well across datasets with heterogeneous distributions.

\subsection{Convergence Performance Comparison}

For a fixed number of communication rounds, we compare the convergence performance of federated TCVAE training.
FedEHR-Gen demonstrates a faster reduction in validation loss across communication rounds than FedAvg as well as its ablated variants, indicating quicker convergence (Appendix~\ref{appendix:comm-efficiency}, Figure~\ref{fig:commu_efficiency}).

\subsection{Ablation Studies}
\label{subsec:ablation}

\begin{figure}[!t]
\centering
\setlength{\abovecaptionskip}{0.1cm}
\includegraphics[width=\linewidth,scale=1.0]{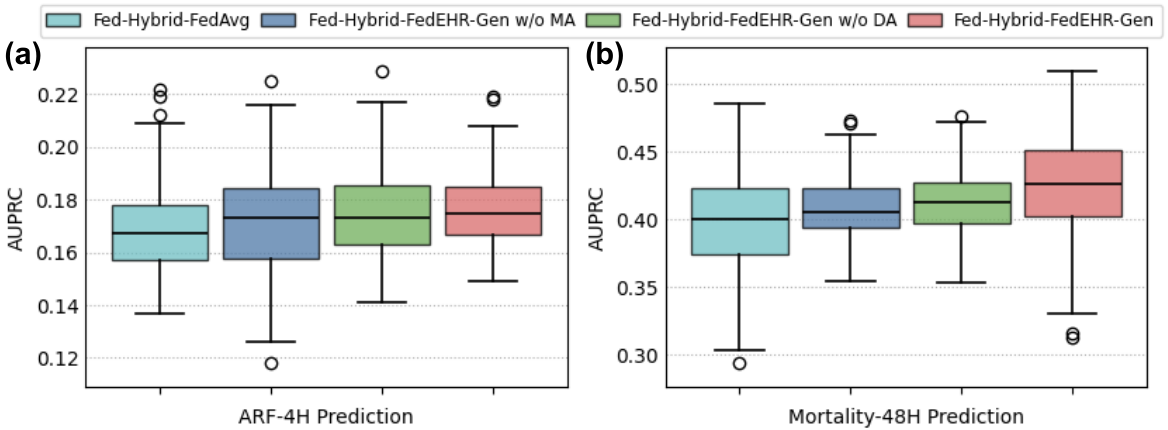}
\caption{Ablation study of FedEHR-Gen showing each component's impact on generation utility (AUPRC).
(a) ARF-4H prediction; (b) Mortality-48H prediction. }
\label{fig:ablation-c5-auprc}
\vspace{-0.3cm}
\end{figure}

We analyze the contributions of \emph{MA} and \emph{DA} by comparing FedEHR-Gen with its ablated variants (Figure~\ref{fig:ablation-c5-auprc}). 
Removing MA (Fed-Hybrid-FedEHR-Gen w/o MA) reduces aggregation to standard parameter averaging for latent encoders, leading to misaligned latent representations across hospitals and a noticeable drop in AUPRC on both ARF-4H and Mortality-48H. Removing DA (Fed-Hybrid-FedEHR-Gen w/o DA) partially mitigates this degradation but remains consistently inferior to the full model, indicating that latent alignment alone is insufficient to handle distributional heterogeneity in federated temporal learning. In contrast, the full Fed-Hybrid-FedEHR-Gen consistently achieves the highest AUPRC, highlighting the complementary roles of MA and DA. Similar trends are also observed in terms of AUROC (Appendix~\ref{appendix:ablation}; Figure~\ref{fig:ablation-c5-auroc}).

\subsection{Adapting Centralized EHR Generation Methods for  Federated Learning as Baselines
}
\label{subsec:more_comparison}
We also consider directly deploying several representative centralized TS-EHR generation models in federated settings. Specifically, we adapt TimeGAN~\cite{yoon2019time}, EHR-M-GAN~\cite{li2023generating}, FlexGen-EHR, and TimeDiff~\cite{tian2024reliable} to the FL paradigm by training them with standard federated optimization, denoted as FedTimeGAN, FedEHR-M-GAN, FedFlexGen-EHR, and FedTimeDiff, respectively.
These methods are not originally designed for federated environments; nevertheless, this setting allows us to examine their behavior and limitations when applied under cross-hospital data isolation.

\begin{figure}[!t]
\centering
\setlength{\abovecaptionskip}{0.1cm}
\includegraphics[width=\linewidth,scale=1.0]{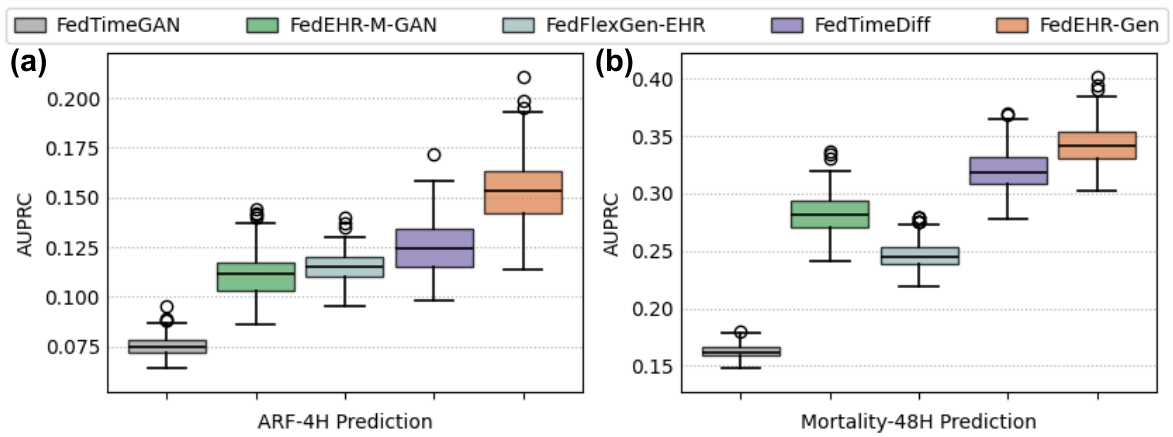}
\caption{Generation of synthetic EHR data generated by four representative centralized time-series generative models when directly applied under federated settings on five eICU hospitals.
(a) AUPRC for ARF-4H prediction; (b) AUPRC for Mortality-48H prediction.}
\label{fig:centralized-gen-in-fl}
\vspace{-0.4cm}
\end{figure}

We compare the downstream utility of synthetic EHR data generated by these federated methods.
Across both ARF-4H and Mortality-48H tasks, these naively federated baselines consistently underperform FedEHR-Gen (Figure~\ref{fig:centralized-gen-in-fl}).
For ARF-4H prediction, FedTimeGAN yields the lowest AUPRC. FedEHR-M-GAN, FedFlexGen-EHR, and FedTimeDiff provide moderate improvements but remain below FedEHR-Gen.
A similar trend is observed for Mortality-48H prediction, where FedEHR-Gen achieves the highest AUPRC among all methods.
These results suggest that centralized TS-EHR generation models, when directly extended to federated settings, have limited ability to preserve task-relevant information under cross-hospital heterogeneity.


\section{Conclusion and Future Work}  \label{sec:conclusion}


This work presents FedEHR-Gen, a federated framework for synthetic time-series EHR generation that enables privacy-preserving collaboration across hospitals.
FedEHR-Gen has several limitations.  The generative model typically requires multiple communication rounds to reach stable convergence, which may introduce additional overhead in large-scale cross-silo federated deployments, and the framework does not provide formal privacy guarantees against strong inference attacks. Future work will focus on improving communication efficiency, investigating stronger privacy mechanisms such as differential privacy, and enhancing robustness to adversarial clients.

\bibliographystyle{ACM-Reference-Format}
\bibliography{kdd}

\clearpage

\appendix

\onecolumn

\begin{center}
  {\LARGE\bfseries Appendix}
\end{center}

\section{Experimental Details}

\subsection{Datasets}
\label{appendix:datasets}

\begin{figure}[b!]
\centering
\includegraphics[width=0.6\linewidth]{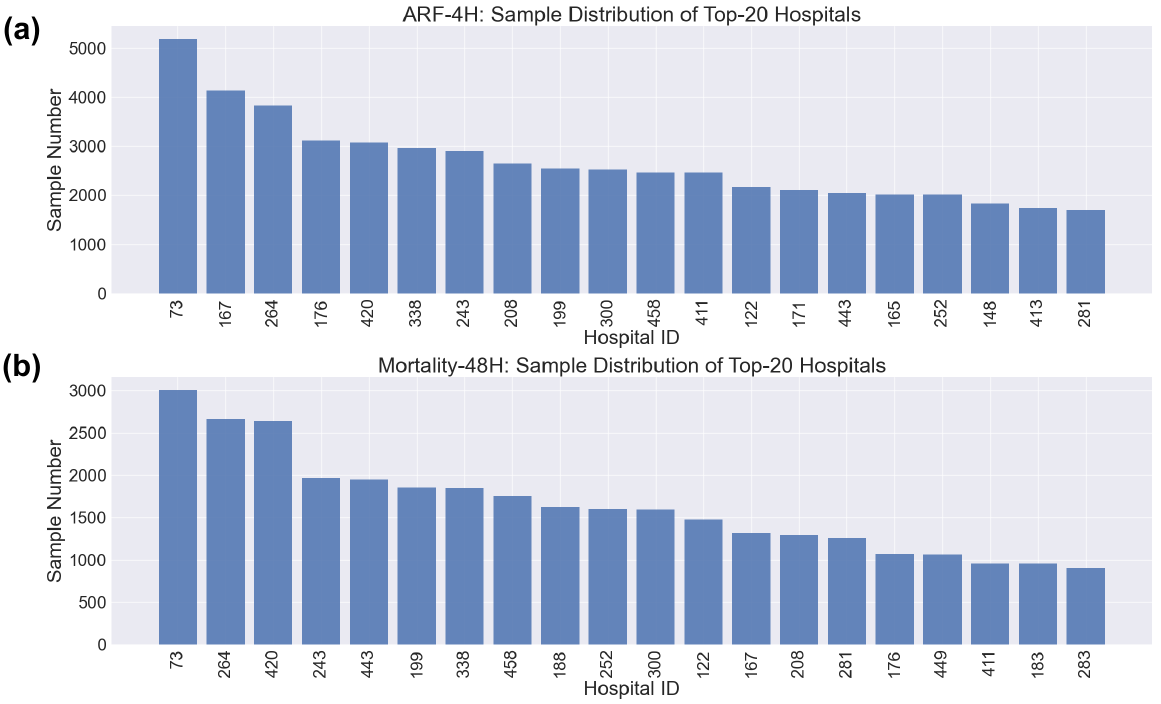}
\caption{Hospital-wise sample distribution for the ARF-4H and Mortality-48H tasks on eICU.
(a) Sample distribution of the top-20 hospitals for ARF-4H prediction.
(b) Sample distribution of the top-20 hospitals for Mortality-48H prediction.}
\label{fig:sample-distribution-over-hips}
\end{figure}

We conduct experiments on two large-scale, real-world intensive care unit (ICU) EHR datasets, MIMIC-III and eICU, both of which contain longitudinal clinical records collected during ICU stays.
In this section, we provide additional details on dataset characteristics, preprocessing, and their suitability for cross-hospital federated evaluation.

\paragraph{\textbf{MIMIC-III}}
MIMIC-III~\cite{mimic3} is a single-center critical care database derived from the Beth Israel
Deaconess Medical Center, containing de-identified EHR data for approximately 60,000 ICU admissions collected between 2001 and 2012. Although MIMIC-III originates from a single institution, it exhibits substantial intra-dataset heterogeneity due to variations in patient demographics, clinical conditions, and treatment practices over time.
In our experiments, MIMIC-III is treated as an independent hospital and is used to evaluate the generalization of federated generative models across institutions.

\paragraph{\textbf{eICU}}
The eICU Collaborative Research Database \cite{eicu} is a large-scale, multi-center ICU
dataset comprising over 200,000 admissions from more than 200 hospitals across the United States, collected between 2014 and 2015. Compared to MIMIC-III, eICU exhibits pronounced cross-hospital heterogeneity, including differences in patient populations, clinical workflows, and data
recording practices. These properties make eICU particularly well-suited for evaluating cross-silo
FL methods under realistic non-IID conditions.

\paragraph{\textbf{Preprocessing and task construction.}}
We adopt the standardized preprocessing pipeline provided by FIDDLE \cite{fiddle}, which converts raw EHR records into discretized multivariate
time-series with fixed temporal resolution.
Clinical events are aggregated into uniform time windows and encoded as high-dimensional, multi-hot binary feature vectors, yielding temporally aligned
sequences suitable for generative modeling.
Based on the processed data, we construct two clinical prediction tasks: in-hospital mortality prediction over a 48-hour window (Mortality-48H) and
short-term acute respiratory failure prediction over a 4-hour window (ARF-4H). The detailed task-specific sample statistics are provided in Table~\ref{tab:dataset_statistics}.

\paragraph{\textbf{Federated hospital partitioning.}}
For eICU, samples are grouped by hospital identifiers to form federated clients in a cross-silo setting. To ensure meaningful client sizes and stable training, we focus on the top 20 hospitals with the largest number of admissions, which together cover the majority of eICU patient records.
The distribution of sample sizes across hospitals indicates substantial variation in data volume, reflecting cross-client heterogeneity (Figure~\ref{fig:sample-distribution-over-hips}). MIMIC-III is treated as an additional standalone client and is not mixed with eICU hospitals during training.

\paragraph{\textbf{Data splits.}}
Within each hospital, we split the cohort into disjoint local training, validation, and test sets with a ratio of 70\%/15\%/15\%.
In addition, we construct a global test set by aggregating local test partitions across hospitals, which is used exclusively for cross-hospital evaluation of
generation fidelity, downstream utility, and privacy risk.

\begin{table}[!t]
\centering
\begin{threeparttable}
\caption{Summary of sample sizes and feature dimensionalities for time-series EHR prediction tasks.}
\label{tab:dataset_statistics}
\fontsize{8}{10}\selectfont
\setlength{\tabcolsep}{6pt}

\begin{tabular}{lccc|ccc}
\toprule
& \multicolumn{3}{c}{MIMIC-III} & \multicolumn{3}{c}{eICU} \\
\cmidrule(lr){2-4} \cmidrule(lr){5-7}
\textbf{Task} & $N$ & $T$ & $D$ & $N$ & $T$ & $D$ \\
\midrule
Mortality-48H & 8{,}577  & 96 & 7{,}307 & 77{,}066  & 146 & 2{,}382 \\
ARF-4H        & 15{,}873 & 98 & 4{,}045 & 138{,}840 & 717 & 5{,}854 \\
\bottomrule
\end{tabular}

\begin{tablenotes}[flushleft]
\footnotesize
\item $N$ denotes the number of samples, $T$ the temporal sequence length, and $D$ the feature dimensionality per time window.
\end{tablenotes}

\end{threeparttable}
\vspace{-0.4cm}
\end{table}

\subsection{Evaluation Metrics} \label{appendix:metrics}

This section provides formal definitions of the evaluation metrics, covering generation fidelity, downstream predictive utility, and privacy risk. 

\paragraph{\textbf{Generation fidelity metrics.}}

\textit{Coefficient of Determination ($R^2$).}
To measure marginal feature-wise fidelity, we compute the coefficient of determination between real and synthetic feature trajectories. Let $\mu^{\mathrm{real}}_{t,d}$ and $\mu^{\mathrm{syn}}_{t,d}$ denote the empirical
means of feature $d$ at time $t$ over real and synthetic datasets, respectively. The $R^2$ score is defined as
\begin{equation}
R^2
=
1
-
\frac{
\sum_{t=1}^{T}\sum_{d=1}^{D}
\left(\mu^{\mathrm{real}}_{t,d}-\mu^{\mathrm{syn}}_{t,d}\right)^2
}{
\sum_{t=1}^{T}\sum_{d=1}^{D}
\left(\mu^{\mathrm{real}}_{t,d}-\bar{\mu}^{\mathrm{real}}\right)^2
},
\end{equation}
where $\bar{\mu}^{\mathrm{real}}$ is the global mean over all real features and time steps. Higher $R^2$ indicates better fidelity.

\textit{Maximum Mean Discrepancy (MMD).}
To assess distributional similarity, we compute the MMD between real and synthetic samples. Given real samples $\{\mathbf{x}_i\}_{i=1}^{N}$ and synthetic samples
$\{\tilde{\mathbf{x}}_j\}_{j=1}^{\tilde{N}}$, MMD is defined as
\begin{equation}
\mathrm{MMD}^2
=
\frac{1}{N^2}\sum_{i,i'} k(\mathbf{x}_i,\mathbf{x}_{i'})
+
\frac{1}{\tilde{N}^2}\sum_{j,j'} k(\tilde{\mathbf{x}}_j,\tilde{\mathbf{x}}_{j'})
-
\frac{2}{N\tilde{N}}\sum_{i,j} k(\mathbf{x}_i,\tilde{\mathbf{x}}_j),
\end{equation}
where $k(\cdot,\cdot)$ is a positive-definite kernel (Gaussian kernel in our experiments). Lower MMD indicates closer distributions.

\textit{Feature-wise Prevalence.} Feature-wise prevalence characterizes the marginal activation frequency of each clinical feature over time and across the patient population.
Let $\mathbf{X}_{n} \in \{0,1\}^{T \times D}$ denote the real EHR time-series sample of patient $n$, and $\tilde{\mathbf{X}}_{n} \in \{0,1\}^{T \times D}$ its synthetic counterpart. The prevalence of the $d$-th feature in the real and synthetic datasets is defined as
\begin{equation}
\pi_d^{\mathrm{real}}
=
\frac{1}{N T}
\sum_{n=1}^{N}
\sum_{t=1}^{T}
x_{n,t,d},
\qquad
\pi_d^{\mathrm{syn}}
=
\frac{1}{\tilde{N} T}
\sum_{n=1}^{\tilde{N}}
\sum_{t=1}^{T}
\tilde{x}_{n,t,d},
\end{equation}
where $N$ and $\tilde{N}$ denote the numbers of real and synthetic
sequences, respectively.
The quantity $\pi_d$ thus represents the empirical probability that
feature $d$ is active at an arbitrary time step across the population.
Comparing $\pi_d^{\mathrm{real}}$ and $\pi_d^{\mathrm{syn}}$ assesses
whether synthetic data preserves the marginal feature distributions of
real EHR time-series, independent of temporal ordering and
cross-feature dependencies.

\paragraph{\textbf{Downstream predictive utility metrics.}}

To evaluate utility, we train downstream clinical prediction models on real, synthetic, or hybrid datasets and test them on held-out real test data.

\textit{Area Under the Receiver Operating Characteristic Curve (AUROC).}
AUROC measures the ranking quality of predicted risk scores across all possible classification thresholds.

\textit{Area Under the Precision--Recall Curve (AUPRC).}
AUPRC is particularly informative for imbalanced clinical tasks and evaluates the trade-off between precision and recall.

\paragraph{\textbf{Privacy risk metrics.}}

\textit{Membership Inference Risk (MIR).}
MIR quantifies the susceptibility of synthetic data to membership inference attacks, i.e., whether an adversary can distinguish training records (members) from non-training records (non-members) given access to the released synthetic dataset. Following an attack-based evaluation protocol, we train a membership inference classifier $a(\cdot)$ that predicts a membership label $m \in \{0,1\}$ for a queried real record $\mathbf{x}$, where $m=1$ indicates that $\mathbf{x}$ was used to train the generator.
The MIR score is computed as the attack advantage:
\begin{equation}
\mathrm{MIR}
=
\Pr\!\bigl(a(\mathbf{x}_{\mathrm{mem}})=1\bigr)
-
\Pr\!\bigl(a(\mathbf{x}_{\mathrm{non}})=1\bigr),
\end{equation}
where $\mathbf{x}_{\mathrm{mem}}$ and $\mathbf{x}_{\mathrm{non}}$ are sampled from the member (training) and non-member (holdout) sets, respectively.
Lower MIR indicates reduced membership leakage (i.e., the attacker performs closer to random guessing).

\textit{Nearest-Neighbor Adversarial Accuracy (NNAA).}
NNAA evaluates whether synthetic samples are overly similar to real records in the data space. For each synthetic sample $\tilde{\mathbf{x}}$, a nearest-neighbor classifier
attempts to distinguish whether its closest neighbor originates from the real or synthetic dataset. NNAA is defined as the adversary's classification accuracy:
\begin{equation}
\mathrm{NNAA}
=
\frac{1}{\tilde{N}}
\sum_{j=1}^{\tilde{N}}
\mathbb{I}\!\left(
\text{NN}(\tilde{\mathbf{x}}_j) \in \mathcal{D}_{\mathrm{real}}
\right),
\end{equation}
where $\mathbb{I}(\cdot)$ is the indicator function. Lower NNAA implies weaker membership distinguishability and better privacy preservation.

\subsection{Implementation Details} \label{appendix:implementation}

\begin{algorithm}[t]
\caption{FedBAE with Matching Aggregation}
\label{alg:fedbae}
\begin{algorithmic}[1]

\INPUT
local EHR datasets $\{\mathbf{X}^{(k)}\}_{k=1}^{K}$;
initial local BAE parameters $\{(\phi^{(k)},\theta^{(k)})\}_{k=1}^{K}$;
sample-size weights $\{\alpha_k\}_{k=1}^{K}$;
communication rounds $R$; local training epochs $E$.

\OUTPUT
semantically aligned global encoder $\phi^{\star}$;
locally adapted decoders $\{\theta^{(k)}\}_{k=1}^{K}$;
aligned latent representations $\{\mathbf{H}^{(k)}\}_{k=1}^{K}$.

\vspace{0.2em}
\FOR{$r = 1$ \TO $R$}

    \STATE $\triangleright$ \textbf{Hospital-side training}
    \FORALL{$k \in \{1,\dots,K\}$ \textbf{in parallel}}
        \STATE Initialize local parameters $(\phi^{(k)},\theta^{(k)}) \leftarrow (\phi^{r-1},\theta^{(k)})$.
        \STATE $\triangleright$ \textbf{Local BAE training}
        \FOR{$e = 1$ \TO $E$}
            \STATE Sample minibatch $\mathbf{X}^{(k)}_{\mathcal{B}} \subset \mathbf{X}^{(k)}$.
            \STATE Compute $\mathbf{Z}=f_{\phi^{(k)}}(\mathbf{X}^{(k)}_{\mathcal{B}})$ using Eq.~(7).
            \STATE Reconstruct $\hat{\mathbf{X}}^{(k)}_{\mathcal{B}}=g_{\theta^{(k)}}(\mathbf{Z})$ using Eq.~(8).
            \STATE Update $(\phi^{(k)},\theta^{(k)})$ by minimizing $\mathcal{L}^{(k)}_{\mathrm{BAE}}$ in Eq.~(9).
            
        \ENDFOR

        \IF{$r > 1$}
            \STATE $\triangleright$ \textbf{Local decoder adaptation}
            \STATE Replace local encoder: $\phi^{(k)} \leftarrow \phi^{r-1}$.
            \STATE Apply the latent-layer permutation to the input channels of the first decoder layer.
            \STATE Freeze encoder parameters $\phi^{(k)}$.
            \STATE Fine-tune decoder $\theta^{(k)}$ by minimizing
            $\mathcal{L}^{(k)}_{\mathrm{BAE}}$.
            \STATE $\triangleright$ \textbf{Local BAE joint fine-tuning}
            \STATE Unfreeze encoder parameters $\phi^{(k)}$.
            \STATE Jointly fine-tune $(\phi^{(k)},\theta^{(k)})$.
        \ENDIF
    
        \STATE Hospital $k$ sends $\phi^{(k)}$ to the server.
    \ENDFOR
    
    \vspace{0.2em}
    \STATE $\triangleright$ \textbf{Server-side matching aggregation}
    \FOR{each encoder hidden layer $\ell = 1,\dots,L$}
        \STATE Estimate permutation matrices $\{\mathbf{\Pi}^{(k)}_{\ell}\}_{k=1}^{K}$ using Eq.~\eqref{eq:matching_obj}.
        \STATE Obtain aggregated layer parameters $(\bar{\mathbf{W}}_{\ell},\bar{\mathbf{b}}_{\ell})$ using Eq.~\eqref{eq:matched_avg}.
    \ENDFOR
    \STATE Form the global encoder $\phi^{r} = \{(\bar{\mathbf{W}}_{\ell},\bar{\mathbf{b}}_{\ell})\}_{\ell=1}^{L}$.
    \vspace{0.2em}

\ENDFOR

$\phi^{\star} \leftarrow \phi^{R}$

\STATE $\triangleright$ \textbf{Construct aligned latent representations per hospital}
\STATE  $\mathbf{H}^{(k)} = f_{\phi^{\star}}(\mathbf{X}^{(k)})$
            using Eq.~\eqref{eq:latent_tensor}.

\STATE \textbf{return} $\phi^{\star}$, $\{\theta^{(k)}\}_{k=1}^{K}$, and $\{\mathbf{H}^{(k)}\}_{k=1}^{K}$.

\end{algorithmic}
\end{algorithm}

\begin{algorithm}[!t]
\caption{FedTCVAE with Distribution-Aware Aggregation}
\label{alg:fedtcvae}
\begin{algorithmic}[1]

\INPUT
aligned latent representations $\{\mathbf{H}^{(k)}\}_{k=1}^{K}$;
initial global parameters $\psi^{0}=\{\psi^{0}_{e},\psi^{0}_{p},\psi^{0}_{d}\}$;
sample-size weights $\{\alpha_k\}_{k=1}^{K}$;
communication rounds $R$; local update steps $E$; temperature $\tau$.

\OUTPUT
global temporal model parameters $\psi^{\star}$.

\vspace{0.2em}
\FOR{$r=1$ \TO $R$}
    \STATE Server broadcasts $\psi^{r-1}$ to all hospitals.
    
    \vspace{0.2em}
    \STATE $\triangleright$ \textbf{Local TCVAE training}
    \FORALL{$k \in \{1,\dots,K\}$ \textbf{in parallel}}
        \STATE Initialize local parameters $\psi^{(k)} \leftarrow \psi^{r-1}$.
        \FOR{$e=1$ \TO $E$}
            \STATE Update recurrent states via Eq.~\eqref{eq:tcvae_rnn}.
            \STATE Compute local ELBO loss $\mathcal{L}^{(k)}_{\mathrm{TCVAE}}$ using Eq.~\eqref{eq:local_tcvae_loss}.
            \STATE Update $\psi^{(k)}$ by minimizing $\mathcal{L}^{(k)}_{\mathrm{TCVAE}}$.
        \ENDFOR
        \STATE Hospital $k$ sends $\psi^{(k)}$ to the server.
    \ENDFOR

    \vspace{0.2em}
    \STATE $\triangleright$ \textbf{Distribution-Aware Aggregation}
    \STATE Compute temporal latent divergences $\{d_{k,j}\}$ using Eq.~\eqref{eq:latent_div}.
    \STATE Compute distribution-aware weights $\{\tilde{\alpha}_k\}$ using Eq.~\eqref{eq:dist_weight}.
    \STATE Aggregate encoder, prior and decoder using Eq.~\eqref{eq:tcvae_agg} to obtain $\psi^{r}_{e}$, $\psi^{r}_{p}$ and $\psi^{r}_{d}$.
    \STATE Set $\psi^{r} \leftarrow \{\psi^{r}_{e}, \psi^{r}_{p}, \psi^{r}_{d}\}$.
\ENDFOR

\STATE \textbf{return} $\psi^{\star} \leftarrow \psi^{R}$.

\end{algorithmic}
\end{algorithm}

\paragraph{\textbf{Federated optimization.}}
Across all experiments, we adopt a local batch size of 512 for client-side training. All models are optimized using Adam with default momentum parameters. For FedBAE, each hospital performs local training until convergence before uploading the encoder parameters to the server. This design facilitates more reliable matching aggregation by ensuring that local latent representations are sufficiently stabilized prior to aggregation. The specific implementation procedure of FedBAE is provided in Algorithm~\ref{alg:fedbae}.

In contrast, for FedTCVAE, we restrict local training to a single epoch per communication round. This choice is motivated by training stability considerations under cross-hospital heterogeneity, and empirically leads to smoother convergence and more stable global optimization. We summarize its implementation details in Algorithm~\ref{alg:fedtcvae}.

\begin{figure}[h!]
\centering
\includegraphics[width=0.6\linewidth]{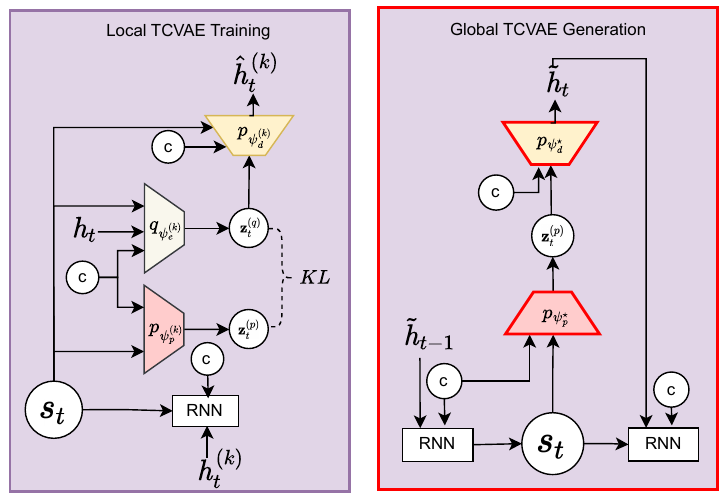}
\caption{Hospital-wise TCVAE training (left) and global TCVAE generation (right).}
\label{fig:tcvae_train_gen}
\end{figure}

\paragraph{\textbf{BAE and TCVAE architectures.}}
The BAE is implemented as a multilayer perceptron with symmetric encoder–decoder architecture. The TCVAE is implemented using a recurrent latent variable model with a two-layer recurrent backbone. Specifically, we employ a two-layer LSTM architecture to capture temporal dependencies in the latent space. The overall training and generation procedures of the TCVAE are illustrated in Figure~\ref{fig:tcvae_train_gen}, with the corresponding client-side training and global generation algorithms summarized in
Algorithm~\ref{alg:local_tcvae_train} and Algorithm~\ref{alg:global_tcvae_gen}, respectively. The detailed network architecture and loss formulation follow the implementation provided in the released codebase.

\begin{algorithm}[!t]
\caption{Local TCVAE training at hospital $k$}
\label{alg:local_tcvae_train}
\begin{algorithmic}[1]
\INPUT
latent sequences $\mathbf{H}^{(k)}=\{\mathbf{h}^{(k)}_{n,1:T}\}_{n=1}^{N_k}$;
conditioning variables $\mathbf{c}$;
initial TCVAE parameters $\psi^{(k)}=\{\psi_e^{(k)},\psi_p^{(k)},\psi_d^{(k)}\}$;
local epochs $E$.
\\
\OUTPUT
updated TCVAE parameters $\psi^{(k)}$.

\vspace{0.2em}
\FOR{$e=1$ \TO $E$}
  \FOR{each mini-batch $\mathcal{B}\subset \mathbf{H}^{(k)}$}
    \STATE Initialize recurrent state $\mathbf{s}_0 \leftarrow \mathbf{0}$.
    \FOR{$t=1$ \TO $T$}
      \STATE Update recurrent state $\mathbf{s}_t$ using Eq.~(17).
      \STATE Compute posterior parameters $(\boldsymbol{\mu}^{(q)}_t,\boldsymbol{\sigma}^{2(q)}_t)$
      and prior parameters $(\boldsymbol{\mu}^{(p)}_t,\boldsymbol{\sigma}^{2(p)}_t)$ using Eq.~(18).
      \STATE Sample $\mathbf{z}_t$ from the posterior via reparameterization:
      $\mathbf{z}_t=\boldsymbol{\mu}^{(q)}_t+\boldsymbol{\sigma}^{(q)}_t\odot\boldsymbol{\epsilon}$,
      $\boldsymbol{\epsilon}\sim\mathcal{N}(\mathbf{0},\mathbf{I})$.
      \STATE Decode latent representation $\hat{\mathbf{h}}_t$ using $p_{\psi_d^{(k)}}(\mathbf{h}_t\mid \mathbf{z}_t,\mathbf{s}_t,\mathbf{c})$.
    \ENDFOR
    \STATE Compute the ELBO loss using Eq.~(\ref{eq:local_tcvae_loss}).
    \STATE Update $\psi^{(k)}$ by SGD.
  \ENDFOR
\ENDFOR
\STATE \textbf{Return} $\psi^{(k)}$.
\end{algorithmic}
\end{algorithm}

\subsection{Reference Model Selection in Matching Aggregation}
\label{appendix:ref-anchor}

In FedBAE matching aggregation (MA), the server aligns each client encoder to a \emph{reference} encoder that defines a canonical neuron ordering before permutation-aware averaging (Eq.~(\ref{eq:matching_obj})--(\ref{eq:matched_avg})). Importantly, the choice of the reference encoder is only required at the \emph{initial matching step}. Once an initial global encoder is obtained, all subsequent communication rounds adopt the \emph{previous-round global encoder} as the reference, thereby maintaining a consistent canonical neuron
ordering across rounds.

\paragraph{Default (FedAvg-based initial anchor).}
At the initial round ($r=0$), we first form a preliminary global encoder by standard sample-size weighted aggregation (FedAvg) over local encoders,
$\phi_{\mathrm{avg}}^{(0)}=\sum_{k=1}^{K}\alpha_k \phi_k^{(0)}$,
with $\alpha_k = N_k / \sum_j N_j$ (Eq.~(\ref{eq:fedavg}).
This aggregated encoder is used as the \emph{initial anchor/reference} to compute layer-wise matchings and obtain the first matched global encoder
$\phi_\star^{(0)}$. For all subsequent rounds ($r \ge 1$), the server directly uses $\phi_\star^{(r-1)}$ as the reference model for MA. This design aligns with the standard FL protocol, where global models evolve
through weighted aggregation and naturally provide a stable canonical ordering.

\paragraph{Alternative (majority-hospital initial anchor).}
We further explore an alternative initialization strategy, where the encoder from the \emph{largest hospital} (i.e., $k^\dagger=\arg\max_k N_k$) is selected as the \emph{initial} anchor/reference model, and all other client encoders are aligned to $\phi_{k^\dagger}^{(0)}$ in the first matching step. After initialization, the same protocol is followed, with the previous-round global encoder serving as the reference in subsequent rounds.

\section{Additional Experiments}

\subsection{Additional Results on Generation Fidelity} \label{appendix:gen-fidelity}

\begin{figure}[h!]
\centering
\includegraphics[width=0.8\linewidth]{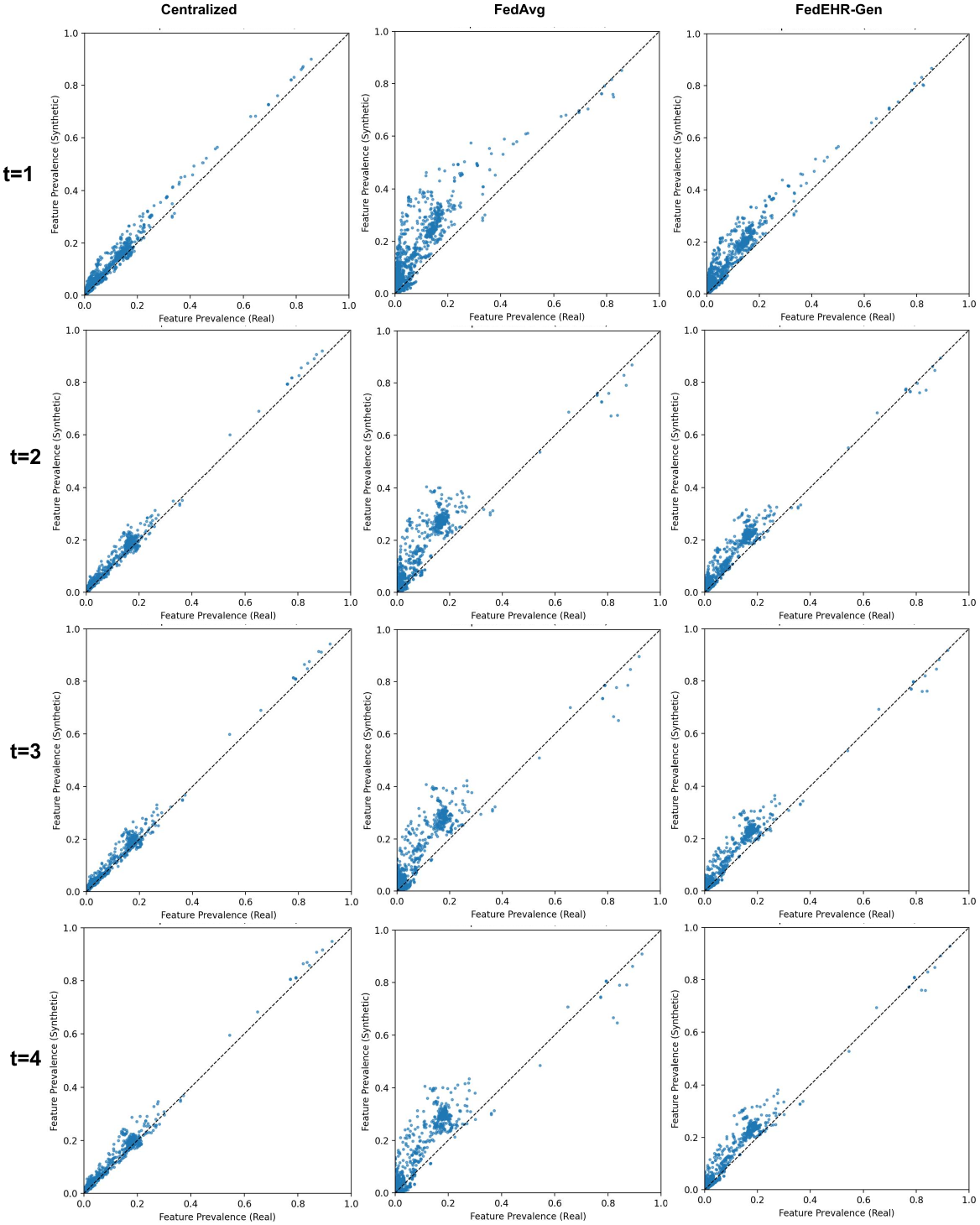}
\caption{Fidelity of generated data in terms of per timestamps feature-wise prevalence on five eICU hospitals for ARF-4H.}
\label{fig:fidelity-arf4h-per-time-feature-prev}
\end{figure}

We further analyze feature-wise prevalence at individual time steps for ARF-4H (Figure~\ref{fig:fidelity-arf4h-per-time-feature-prev}). Each
row corresponds to one timestamp. While centralized training closely follows the diagonal across all timestamps, FedAvg
exhibits larger deviations, particularly for low-prevalence features. FedEHR-Gen consistently yields tighter alignment at each timestamp, indicating more stable preservation of feature-wise prevalence over time.

\begin{figure}[!t]
\centering
\includegraphics[width=0.9\linewidth]{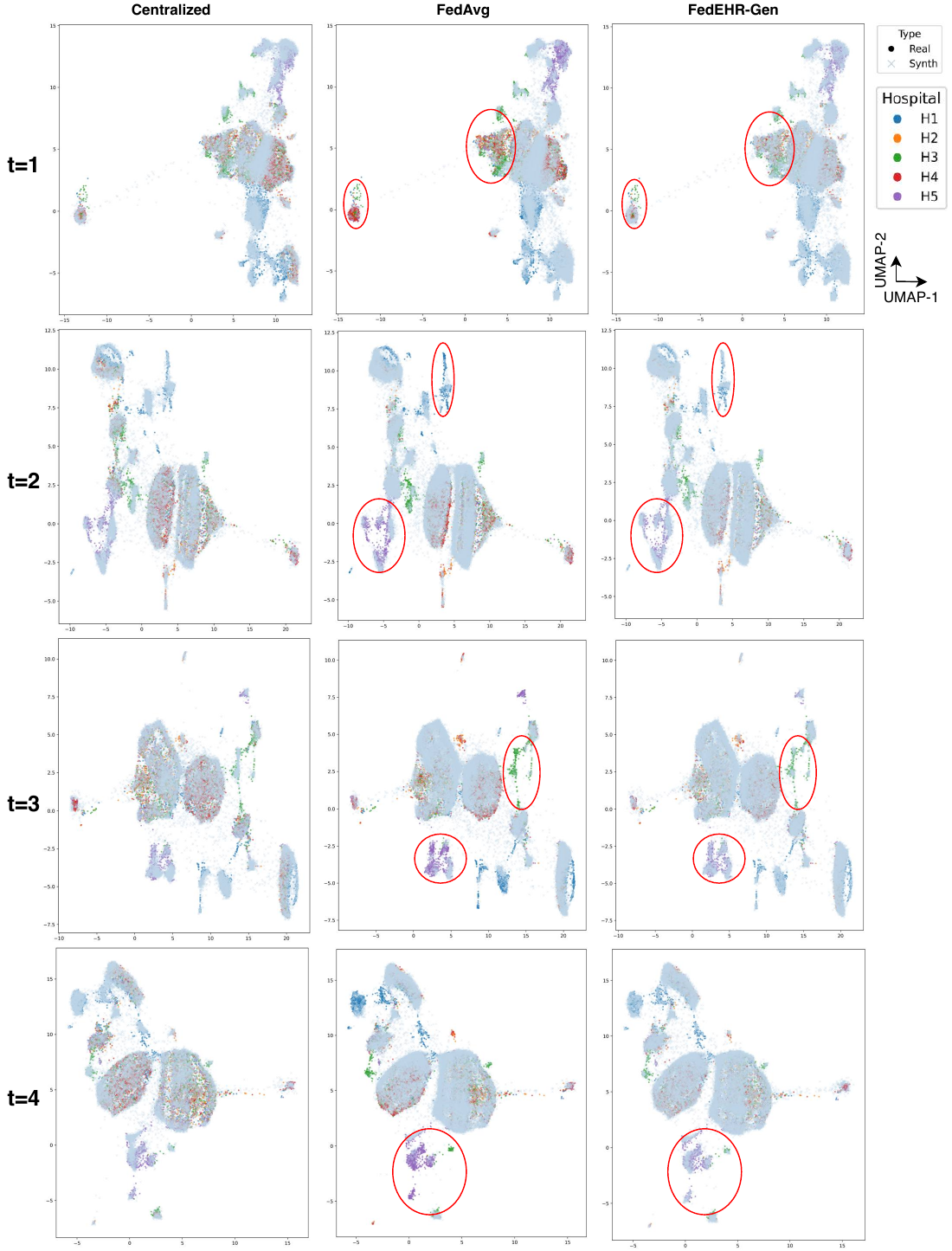}
\caption{Fidelity of generated data in terms of per timestamps UMAP visualization on five eICU hospitals for ARF-4H.}
\label{fig:fidelity-arf4h-per-time-umap}
\end{figure}

We further analyze per-timestamp UMAP visualizations for ARF-4H (Fig.~\ref{fig:fidelity-arf4h-per-time-umap}). Each row corresponds to a single timestamp. While the centralized baseline preserves broad coverage of samples from all hospitals across timestamps, FedAvg exhibits clear coverage collapse at multiple time steps, where samples from certain hospitals are partially missing or collapse into sparse regions of the embedding space. This observation indicates mode dropping and reduced distributional diversity under naive federated aggregation. In contrast, FedEHR-Gen consistently maintains broader and more balanced coverage across hospitals at each timestamp, yielding more diverse and complete synthetic distributions that better align with the real data manifold.

\begin{algorithm}[!t]
\caption{Global TCVAE Generation}
\label{alg:global_tcvae_gen}
\begin{algorithmic}[1]
\INPUT
well trained global TCVAE parameters $\psi^\star=\{\psi_p^\star,\psi_d^\star\}$;
conditioning variables $\mathbf{c}$; sequence length $T$.
\\
\OUTPUT
synthetic latent sequence $\tilde{\mathbf{h}}_{1:T}$.

\STATE Initialize $\mathbf{s}_0\leftarrow \mathbf{0}$ and $\tilde{\mathbf{h}}_0\leftarrow \mathbf{0}$.
\FOR{$t=1$ \TO $T$}
  \STATE Update recurrent state $\mathbf{s}_t$ using Eq.~\eqref{eq:tcvae_rnn}.
  \STATE Compute prior parameters $(\boldsymbol{\mu}^{(p)}_t, \boldsymbol{\sigma}^{2(p)}_t)$ using $p_{\psi_p^\star}(\mathbf{z}_t\mid \mathbf{s}_t,\mathbf{c})$.
\STATE Sample latent variable via reparameterization: \\
$\tilde{\mathbf{z}}_t=\boldsymbol{\mu}^{(p)}_t+\boldsymbol{\sigma}^{(p)}_t\odot\boldsymbol{\epsilon},
\ \boldsymbol{\epsilon}\sim\mathcal{N}(\mathbf{0},\mathbf{I})$.
  \STATE Generate latent representation $\tilde{\mathbf{h}}_t$ using $p_{\psi_d^\star}(\mathbf{h}_t\mid \tilde{\mathbf{z}}_t,\mathbf{s}_t,\mathbf{c})$.
\ENDFOR
\STATE \textbf{Return} $\tilde{\mathbf{h}}_{1:T}$.
\end{algorithmic}
\end{algorithm}

\subsection{Additional Results on Generation Utility} \label{appendix:utility}

\begin{figure}[!t]
\centering
\setlength{\abovecaptionskip}{0.1cm}
\includegraphics[width=\linewidth,scale=1.0]{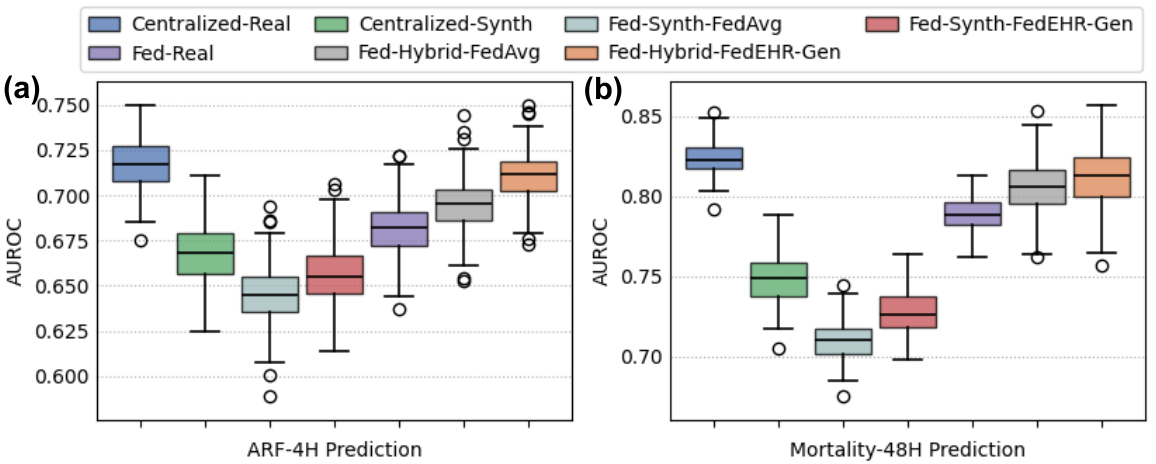}
\caption{Prediction performance on the clinical tasks with five eICU hospitals. (a) AUROC for ARF-4H prediction; (b) AUROC for Mortality-48H prediction.}
\label{fig:fig:utility_synth_hybrid_c5_auroc}
\end{figure}

Generation utility varies systematically across training strategies, with models trained on real data achieving the highest AUROC and serving as a clear upper bound for downstream prediction performance (\autoref{fig:fig:utility_synth_hybrid_c5_auroc}). In contrast, classifiers trained solely on synthetic data exhibit consistently lower AUROC across both tasks, indicating that synthetic EHRs cannot fully substitute real data for downstream prediction. Incorporating synthetic samples as a complementary signal in hybrid training leads to improved performance over both Fed-Real and Fed-Synth settings, suggesting that synthetic data is most effective when used to augment, rather than replace, real observations. Among all federated approaches, the Fed-Hybrid-FedEHR-Gen strategy attains the highest AUROC for both ARF-4H and Mortality-48H prediction, reflecting the advantage of higher-fidelity synthetic data in enhancing downstream utility under federated constraints (\autoref{fig:fig:utility_synth_hybrid_c5_auroc}).

\subsection{Additional Results on Scalability} \label{appendix:scalability}

\begin{figure*}[!t]
\centering
\setlength{\abovecaptionskip}{0.1cm}
\includegraphics[width=0.9\linewidth,scale=1.0]{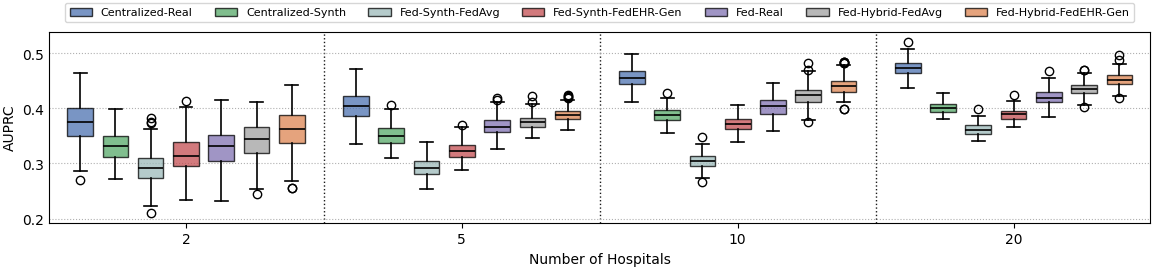}
\caption{Effect of varying federated scales (number of hospitals) on Mortality-48H AUPRC on eICU across seven methods.
}
\label{fig:c5-scale-auprc-mortality}
\end{figure*}

For the Mortality-48H task, AUPRC increases steadily as the number of participating hospitals grows from 2 to 20, indicating improved predictive performance with larger federated scales (Figure~\ref{fig:c5-scale-auprc-mortality}). 
Hybrid federated training consistently outperforms real-only and synthetic-only baselines across all scales, demonstrating the effectiveness of combining real and synthetic data. 
Among federated methods, \textit{Fed-Hybrid-FedEHR-Gen} achieves the highest AUPRC throughout, confirming its robustness and scalability in large cross-hospital settings (Figure~\ref{fig:c5-scale-auprc-mortality}).

Generation fidelity generally degrades as the number of hospitals increases, reflected by a consistent decrease in $R^2$ across all methods and an overall elevation of MMD under larger federation scales (Table~\ref{tab:generation_fidelity_varying_hosp}). 
For centralized training, $R^2$ decreases from 0.821 at 2 hospitals to 0.693 at 20 hospitals, while MMD increases from 0.742 to 0.887, indicating growing difficulty in preserving feature-wise distributions under stronger cross-hospital heterogeneity.
Among federated approaches, FedEHR-Gen consistently outperforms FedAvg across all settings, achieving higher $R^2$ and lower MMD at each federation scale. For example, at 10 hospitals, FedEHR-Gen improves $R^2$ from 0.598 to 0.672 and reduces MMD from 0.896 to 0.869 compared with FedAvg; similar advantages are observed at 20 hospitals, where $R^2$ increases from 0.551 to 0.648 and MMD decreases from 0.915 to 0.874. These results indicate that FedEHR-Gen achieves more accurate alignment between synthetic and real data distributions and exhibits improved robustness under large-scale cross-hospital federated settings.

\begin{table*}[!h]
\centering
\caption{\textbf{Generation Fidelity under varying numbers of hospitals.}
$R^2$ ($\uparrow$) and MMD ($\downarrow$) between synthetic and real datasets for ARF-4H under different federation scales.}
\label{tab:generation_fidelity_varying_hosp}
\fontsize{8}{10}\selectfont
\renewcommand{\arraystretch}{1.0}
\begin{tabular}{lcccccccc}
\toprule
\multirow{2}{*}{Method} 
& \multicolumn{2}{c}{ARF-4H (2 Hosp.)} 
& \multicolumn{2}{c}{ARF-4H (5 Hosp.)} 
& \multicolumn{2}{c}{ARF-4H (10 Hosp.)} 
& \multicolumn{2}{c}{ARF-4H (20 Hosp.)} \\
\cmidrule(lr){2-3} \cmidrule(lr){4-5} \cmidrule(lr){6-7} \cmidrule(lr){8-9}
& $R^2$ ($\uparrow$) & MMD ($\downarrow$)
& $R^2$ ($\uparrow$) & MMD ($\downarrow$)
& $R^2$ ($\uparrow$) & MMD ($\downarrow$)
& $R^2$ ($\uparrow$) & MMD ($\downarrow$) \\
\midrule
Centralized 
& $0.821 \pm 0.030$ & $0.742 \pm 0.030$
& $0.768 \pm 0.038$ & $0.842 \pm 0.020$
& $0.716 \pm 0.039$ & $0.853 \pm 0.025$
& $0.693 \pm 0.041$ & $0.887 \pm 0.028$ \\

FedAvg 
& $0.715 \pm 0.037$ & $0.813 \pm 0.031$
& $0.672 \pm 0.041$ & $0.892 \pm 0.021$
& $0.598 \pm 0.036$ & $0.896 \pm 0.047$
& $0.551 \pm 0.040$ & $0.915 \pm 0.021$ \\

\textbf{FedEHR-Gen}
& $\mathbf{0.788 \pm 0.034}$ & $\mathbf{0.776 \pm 0.024}$
& $\mathbf{0.721 \pm 0.040}$ & $\mathbf{0.863 \pm 0.024}$
& $\mathbf{0.672 \pm 0.040}$ & $\mathbf{0.869 \pm 0.029}$
& $\mathbf{0.648 \pm 0.046}$ & $\mathbf{0.874 \pm 0.022}$ \\
\bottomrule
\end{tabular}
\end{table*}

\begin{figure*}[!t]
\centering
\setlength{\abovecaptionskip}{0.1cm}
\includegraphics[width=\linewidth,scale=1.0]{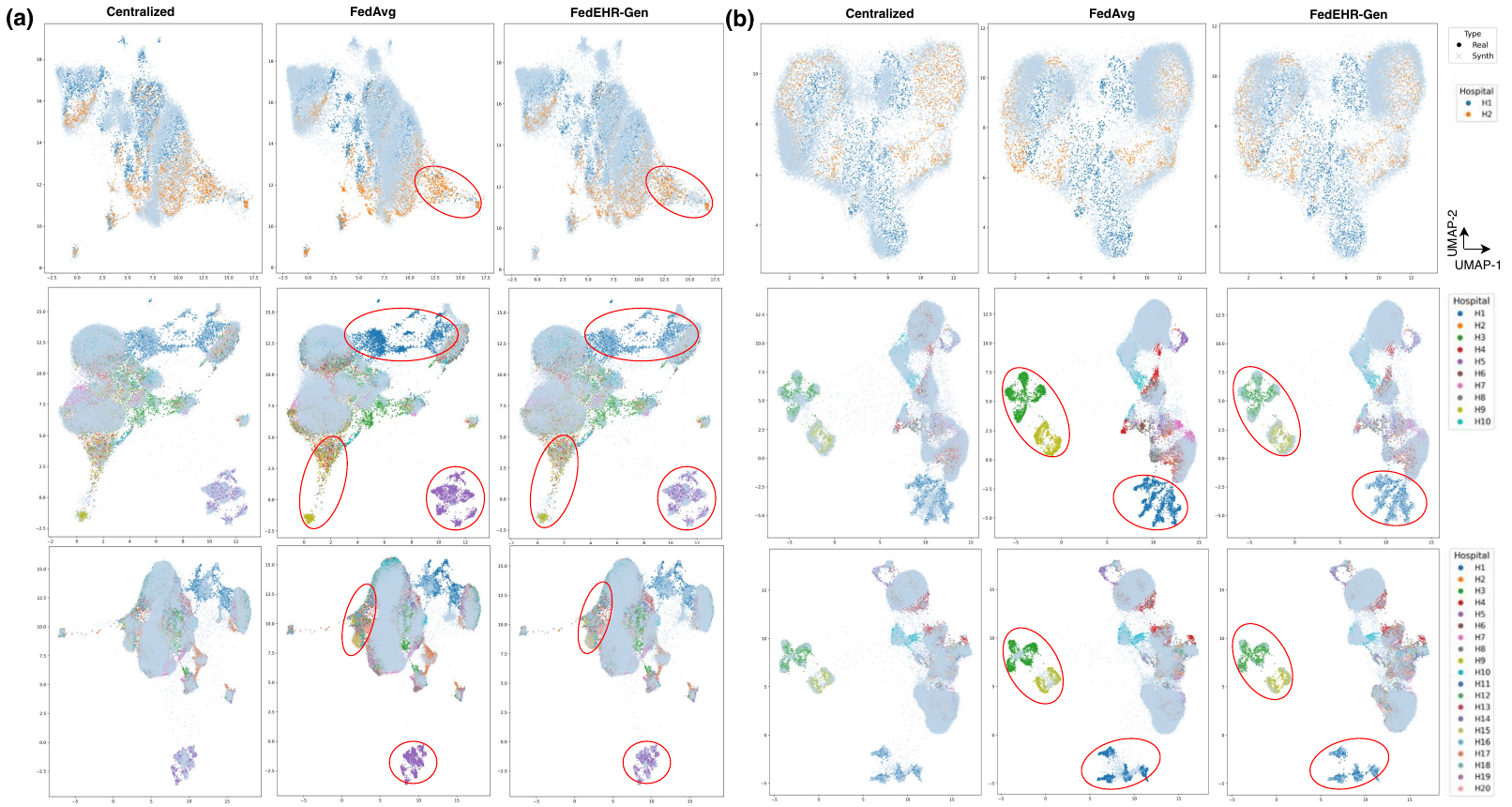}
\caption{Generation fidelity of UMAP visualization on eICU with varying hospitals. (a) ARF-4H; (b) Mortality-48H.}
\label{fig:fidelity-varying-hospitals-umap}
\end{figure*}

We also provides UMAP visualizations of real and synthetic samples under different federated scales (2, 10, and 20 hospitals) for three generation strategies (Figure~\ref{fig:fidelity-varying-hospitals-umap}).
The centralized baseline consistently achieves strong alignment between real and synthetic data across all scales. As the number of hospitals increases, FedAvg suffers from reduced coverage in the embedding space, with synthetic samples from some hospitals becoming under-represented or missing, as indicated by the red circled regions.
In contrast, FedEHR-Gen preserves more comprehensive and balanced coverage across hospitals, yielding synthetic distributions that remain well aligned with the real data manifold under varying federated scales (Figure~\ref{fig:fidelity-varying-hospitals-umap}).

\subsection{Additional Results on Cross-Dataset Generation} \label{appendix:cross-dataset}

\begin{figure}[!t]
\centering
\setlength{\abovecaptionskip}{0.1cm}
\includegraphics[width=0.95\linewidth,scale=1.0]{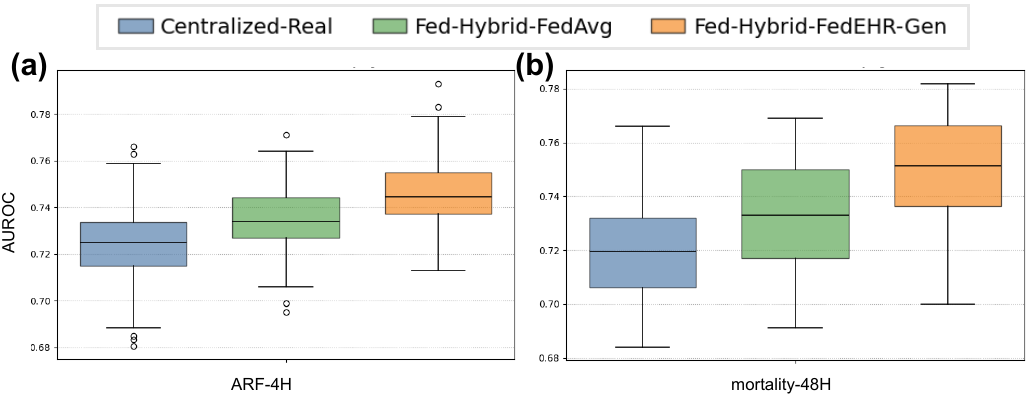}
\caption{Cross-dataset generation utility on MIMIC-III. AUROC for models trained with synthetic data generated by a generator trained on eICU and evaluated on MIMIC-III.
(a) ARF-4H prediction; (b) Mortality-48H prediction.}
\label{fig:cross-datasets-c5-auroc}
\end{figure}

When evaluated using AUROC, we observe consistent trends with those reported for AUPRC. Hybrid training on MIMIC-III improves predictive performance over real-only federated baselines across both ARF-4H and Mortality-48H tasks (Figure~\ref{fig:cross-datasets-c5-auroc}). In particular, Fed-Hybrid-FedEHR-Gen achieves the highest AUROC among all methods, outperforming the FedAvg-based hybrid baseline.
These results further confirm that synthetic data generated by the proposed federated generator provides complementary predictive signals and improves cross-dataset generalization under domain shift (Figure~\ref{fig:cross-datasets-c5-auroc}).

\subsection{Convergence Comparison for Federated TCVAE Training} \label{appendix:comm-efficiency}

\begin{figure}[!t]
\centering
\setlength{\abovecaptionskip}{0.1cm}
\includegraphics[width=0.8\linewidth,scale=1.0]{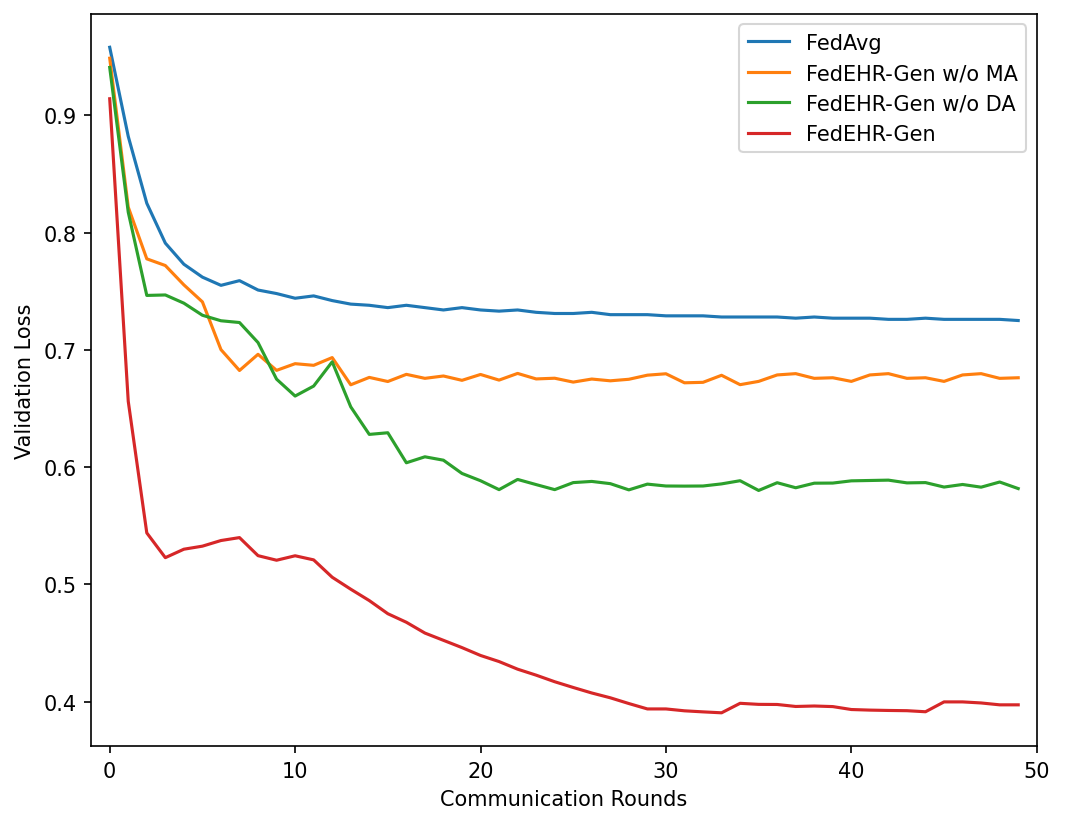}
\caption{Communication efficiency comparison between FedAvg, FedEHR-Gen, and its ablated variants for TCVAE training on five eICU hospitals for ARF-4H.}
\label{fig:commu_efficiency}
\end{figure}

The validation loss decreases steadily with increasing communication rounds for all methods, indicating progressive convergence during federated TCVAE training (\autoref{fig:commu_efficiency}). Compared with FedAvg, FedEHR-Gen exhibits a consistently faster convergence rate and achieves lower validation loss throughout the training process, reflecting improved communication efficiency. Removing either the MA or the DA leads to slower convergence and higher final loss, suggesting that both components contribute to stabilizing optimization under federated settings. Notably, FedEHR-Gen reaches a comparable validation loss with substantially fewer communication rounds, highlighting its advantage in reducing communication overhead while maintaining effective training dynamics (\autoref{fig:commu_efficiency}).

\subsection{Additional Results on Ablation Studies} \label{appendix:ablation}

\begin{figure}[!t]
\centering
\setlength{\abovecaptionskip}{0.1cm}
\includegraphics[width=\linewidth,scale=1.0]{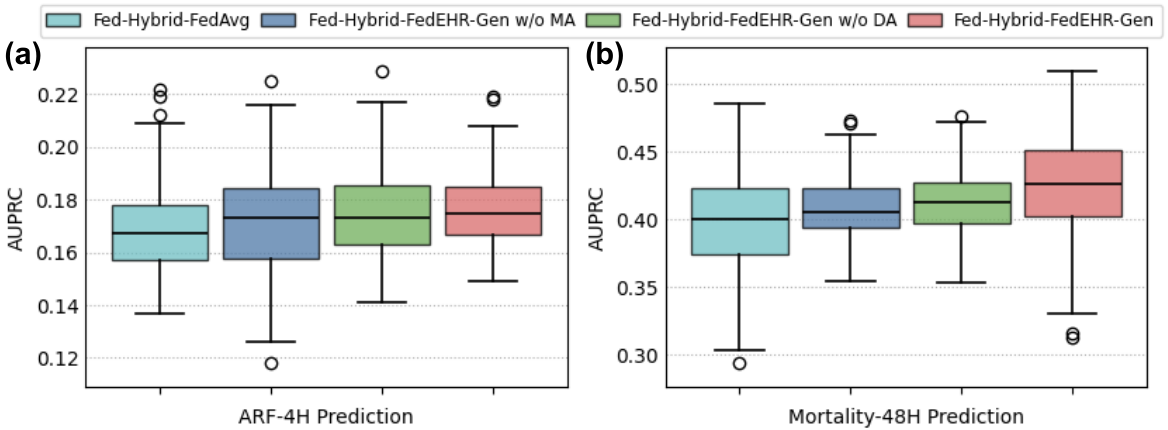}
\caption{Ablation study of FedEHR-Gen.
Effect of different model components on generation utility measured by downstream AUROC.
(a) ARF-4H prediction; (b) Mortality-48H prediction. }
\label{fig:ablation-c5-auroc}
\vspace{-0.4cm}
\end{figure}

Consistent trends are observed when evaluated using AUROC.
Removing MA (Fed-Hybrid-FedEHR-Gen w/o MA) leads to a clear performance degradation on both ARF-4H and Mortality-48H, while removing DA (Fed-Hybrid-FedEHR-Gen w/o DA) results in a milder but consistent drop.
In contrast, the full Fed-Hybrid-FedEHR-Gen model achieves the highest AUROC across both tasks, highlighting the complementary contributions of MA and DA (Figure~\ref{fig:ablation-c5-auroc}).

\section{Related Work} \label{appendix:related_work}


\textbf{\textit{Federated Learning}} (FL) enables collaborative model training across institutions while keeping sensitive data local, offering a principled foundation for privacy-preserving healthcare analytics. The vanilla FedAvg algorithm~\cite{fedavg} and subsequent methods designed to mitigate optimization drift under non-IID data, such as FedProx~\cite{fedprox}, SCAFFOLD~\cite{Scaffold}, FedDyn~\cite{feddyn}, and MOON~\cite{moon}, demonstrate that distributional heterogeneity remains the dominant obstacle in practical FL deployments. This challenge is especially pronounced in Electronic Health Records (EHR), where hospitals differ widely in patient populations, clinical workflows, coding practices, and temporal measurement patterns. Motivated by these constraints, a growing body of research has applied FL to multi-hospital EHR modeling, spanning tasks such as federated risk prediction ~\cite{fedweight, dayan2021federated, al2025fedcomdist}, phenotype classification \cite{dang2022federated}, and large-scale clinical representation learning \cite{liu2019two, zhou2025representation}. While these studies confirm the feasibility of privacy-preserving collaboration across institutions, they consistently report substantial performance degradation caused by severe cross-hospital heterogeneity, manifesting in representation misalignment, unstable training dynamics, and reduced generalization across clinical sites.

\textbf{\textit{Synthetic EHR generation}} has been extensively studied in centralized settings as a means to alleviate data scarcity and enable safe clinical model development. Existing approaches span a broad range of generative paradigms, including GAN-based models for structured medical records (e.g., MedGAN~\cite{choi2017generating}, GRACE~\cite{grace}), VAE-based frameworks that capture latent clinical uncertainty (e.g., EMR-VAE~\cite{theodorou2023synthesize}), and sequence-aware generators for temporal EHR, such as recurrent or adversarial time-series models (e.g., TimeGAN~\cite{yoon2019time}, RCGAN~\cite{esteban2017real}). More recent work further explores advanced time-series generation techniques, including  diffusion-based models (e.g., TimeDiff~\cite{tian2024reliable}, TarDiff~\cite{deng2025tardiff}), which achieve high-fidelity temporal synthesis in centralized environments. While these methods demonstrate strong generation quality and downstream utility, they universally assume centralized access to patient records and do not account for cross-institution heterogeneity, rendering them unsuitable for privacy-restricted, multi-hospital collaboration scenarios.


\textbf{\textit{Federated generative modeling}} has been actively studied in recent years, with representative work on federated GANs~\cite{han2025fed, ma2023flgan}, VAEs~\cite{esposito2025transfer, duan2023federated}, and diffusion-based models~\cite{stanley2024phoenix, li2024feddiff, peng2025federated}. These approaches demonstrate the feasibility of training generative models without centralizing sensitive data, but are primarily developed for dense and continuous modalities such as images or text. Clinical EHR, by contrast, are characterized by extreme sparsity, multi-hot encoding, and severe cross-hospital heterogeneity, which are not explicitly addressed by existing federated generative frameworks. As a result, directly applying prior methods to multi-hospital time-series EHR remains challenging and motivates the need for domain-specific federated generative modeling.

\end{document}